\documentclass[preprint]{elsarticle} 

\makeatletter
\def\ps@pprintTitle{%
\let\@oddhead\@empty
\let\@evenhead\@empty
\def\@oddfoot{\centerline{\thepage}}%
\let\@evenfoot\@oddfoot}
\makeatother

\usepackage{etoolbox}
\patchcmd{\MaketitleBox}{\footnotesize\itshape\elsaddress\par\vskip36pt}{\footnotesize\itshape\elsaddress\par\parbox[b][36pt]{\linewidth}{\vfill\hfill\textnormal{\today}\hfill\null\vfill}}{}{}%
\patchcmd{\pprintMaketitle}{\footnotesize\itshape\elsaddress\par\vskip36pt}{\footnotesize\itshape\elsaddress\par\parbox[b][36pt]{\linewidth}{\vfill\hfill\textnormal{\today}\hfill\null\vfill}}{}{}%

\usepackage[
colorlinks=true,%
breaklinks=true,%
linkcolor=blue,
urlcolor=blue,%
citecolor=blue,%
pdftitle={Title of paper}, 
pdfkeywords={Keywords},	
pdfauthor={Authors},		
bookmarksopen=false,
pdfpagemode=UseNone]{hyperref}
\usepackage{algorithm}
\usepackage{algorithmicx}
\usepackage{algpseudocode}
\usepackage{tabularx}
\usepackage{array}
\usepackage{multirow}
\usepackage{threeparttable}
\usepackage{booktabs}
\usepackage{graphicx}
\usepackage{arydshln}
\usepackage{setspace}   
\usepackage{xcolor}
\usepackage[mathlines, running]{lineno}
\usepackage{epstopdf}
\usepackage{mathrsfs}
\usepackage{amsfonts}
\usepackage{tikz}
\usepackage{longtable}
\usepackage{caption}
\usepackage{makecell}

\makeatletter

\usepackage[margin = 1.2in]{geometry}
\usepackage[T1]{fontenc}
\usepackage[english]{babel}
\usepackage[latin1]{inputenc}
\usepackage{mathtools}
\usepackage{amsmath}
\usepackage{amsfonts}
\usepackage{amsthm}
\usepackage{amssymb}
\usepackage{array}
\usepackage{subcaption}
\usepackage{siunitx}

\usepackage{color}
\usepackage{url}
\usepackage{cleveref}
\usepackage{graphicx}
\usepackage{breakcites}
\usepackage{soul} 
\usepackage{enumitem}
\usepackage[title,titletoc,toc]{appendix}

\usetikzlibrary{shapes.geometric, arrows}

\tikzstyle{startstop} = [rectangle, rounded corners, minimum width=3cm, minimum height=1cm,text centered, draw=black, dashed]
\tikzstyle{arrow} = [thick,->,>=stealth]

\usepackage[textsize=tiny]{todonotes}
 
\newcommand{\RNum}[1]{\uppercase\expandafter{\romannumeral #1\relax}}
\usepackage{amsthm}
\makeatletter
\def\els@aparagraph[#1]#2{\elsparagraph[#1]{#2\@addpunct{.}}}
\def\els@bparagraph#1{\elsparagraph*{#1\@addpunct{.}}}
\makeatother
\usepackage{amsmath}
\DeclareMathOperator*{\argmax}{arg\,max}

\newtheorem{problem}{Problem}

{\noindent {\textbf{Proof}:} }%
{\hfill $\Box$ \\[1ex] }

\newcommand{\bit}{\begin{itemize}}
\newcommand{\eit}{\end{itemize}}
\newcommand{\ben}{\begin{enumerate}}
\newcommand{\een}{\end{enumerate}}

\newcommand {\real} {\mathbb{R}}
\newcommand {\nat} {\mathbb{N}}

\newcommand{\rd}{\text{\upshape d}} 

\newcommand{\bX}{\ensuremath{\mathbf{X}}}

\newcommand{\bx}{\ensuremath{\mathbf{x}}}

\newcommand{\cB}{\ensuremath{\mathcal{B}}}

\newcommand{\cF}{\ensuremath{\mathcal{F}}}

\graphicspath{{./figures/}}
\begin{document}	
\begin{frontmatter}
		
\title{Spline Dimensional Decomposition with Interpolation-based Optimal Knot Selection for Stochastic Dynamic Analysis}
		
		
\author[hyu]{Yeonsu Kim}
\ead{lelghan@hanyang.ac.kr}
\author[hyu]{Junhan Lee}
\author[ucsd]{Bingran Wang}
\ead{b6wang@ucsd.edu}
\author[ucsd]{John T. Hwang}
\ead{jhwang@ucsd.edu}
\author[hyu]{Dongjin Lee\corref{cor2}}
\ead{dlee46@hanyang.ac.kr}
		
\cortext[cor2]{Corresponding author}
		
\address[hyu]{Department of Automotive Engineering, Hanyang University, Seoul, Republic of Korea}
\address[ucsd]{Department of Mechanical and Aerospace Engineering, University of California San Diego, CA, USA}


\begin{abstract}
Forward uncertainty quantification in dynamical systems is challenging due to non-smooth or locally oscillating nonlinear behaviors. 
Spline dimensional decomposition (SDD) effectively addresses such nonlinearity by partitioning input coordinates via knot placement; yet, its accuracy is highly sensitive to the location of internal knots. 
%
%
We propose a computationally efficient, interpolation-based method for optimal knot selection in SDD. 
The method involves three steps: (1) interpolating input-output profiles, (2) defining subinterval-based reference regions, and (3) selecting optimal knot locations at maximum gradient points within each region. 
The resulting knot vector is then applied to SDD for accurate approximation of non-smooth and locally oscillating responses. 
A modal analysis of a lower control arm demonstrates that SDD with the proposed knot selection achieves higher accuracy than SDD with uniformly or randomly spaced knots, and also a Gaussian process surrogate model.  
In this example, the proposed SDD exhibits the lowest relative variance error (2.89\%) in the first natural frequency distribution, compared to SDD with uniformly spaced knots (12.310\%), randomly spaced knots (15.274\%), and the Gaussian process (5.319\%). 
%
%
The scalability and applicability of proposed method are demonstrated through stochastic and reliability analyses of one- and three-dimensional mathematical functions, along with a ten-dimensional lower control arm model. 
The results confirm that both second-moment statistics and reliability estimates can be accurately achieved with only a few hundred function evaluations or finite element simulations. 
\end{abstract}

\begin{keyword}
 Uncertainty quantification, Surrogate model, Optimal knot vector, Spline dimensional decomposition, Dynamical system, Stochastic properties, Reliability analysis
\end{keyword}

\end{frontmatter}
\section{Introduction} \label{sec:intro}
Uncertainty quantification (UQ) plays a pivotal role in the design and analysis of reliable engineering systems, as real-world input parameters---such as material properties, geometric dimensions, boundary conditions, and loading forces---are subject to inherent variability and randomness. 
Forward UQ propagates uncertainties from the input space to the output space. This process enables the estimation of key statistical properties such as the mean, variance, confidence intervals, and tail probabilities (e.g., failure probability) to ensure system safety and requirements under uncertain conditions. 
A common approach to forward UQ is the use of crude Monte Carlo simulation (MCS), wherein a large number of independent realizations of input variables are generated based on their prescribed probability distributions, and the corresponding system responses are computed. 
While MCS is conceptually straightforward and asymptotically accurate, it suffers from extremely slow convergence rates, particularly when estimating low-probability events or when the computational model is expensive to evaluate. Consequently, the method becomes computationally inefficient or even infeasible when applied to high-fidelity engineering models~\cite{zhang2021modern,kaintura2018review}.


Surrogate modeling has emerged as a powerful strategy to mitigate the high computational cost of UQ, especially in scenarios involving complex or high-fidelity simulation models. Surrogate models approximate the input-output relationship using a limited number of simulation samples, thereby enabling efficient propagation of uncertainty with significantly reduced computational effort~\cite{wang2022recent,marrel2024probabilistic}. A range of state-of-the-art surrogate modeling techniques have been developed for UQ applications, including artificial neural networks (ANNs)~\cite{zhou2023recent}, polynomial chaos expansion (PCE)~\cite{lee2020practical}, polynomial dimensional decomposition (PDD)~\cite{rahman2008polynomial}, spline chaos expansion (SCE)~\cite{rahman2020spline}, spline dimensional decomposition (SDD)~\cite{rahman2022spline}, reduced order modeling~\cite{kadeethum2023epistemic}, Kriging~\cite{kasdorf2024kriging} and more~\cite{liu2022hybrid,lee2023multifidelity,zhao2025adaptive}. Each method offers unique advantages in terms of scalability, accuracy, and suitability for various types of uncertainty and system behavior. Among these, SDD particularly excels at capturing strong nonlinearity and both local and global nonsmoothness in the response surface. SDD constructs the surrogate using hierarchical spline functions combined with ANOVA decomposition, allowing it to efficiently capture discontinuities and steep gradients in high-dimensional input spaces. 

The SDD uses B-splines as basis functions in the surrogate model~\cite{rahman2020spline} to handle non-smooth regions. B-spline knot vectors divide the domain into intervals, applying polynomial functions of specific orders to capture localized variations precisely. This approach makes B-spline-based surrogate models especially effective for accurately predicting nonlinear responses---such as harmonic or transient dynamic responses---exhibiting nonsmooth or non-differentiable characteristics in dynamic engineering systems~\cite{rahman2021orthogonal}. 
However, the accuracy of SDD relies heavily on the placement of knots as these affect the model's ability to handle regions with sparse data or sharp gradients~\cite{dixler2021uncertainty,spiriti2013knot}. As a result, the development of algorithms that adjust a proper placement of internal knots has enabled the construction of adaptive B-spline based models~\cite{rehme2021b,vuillod2024handling,dertimanis2018data}.
A recent study~\cite{rehme2021b} proposes a modified Not-a-Knot condition that enforces zero second derivatives at boundary points for B-splines, incorporating hierarchical structures with spatial adaptivity to compute expansion coefficients efficiently.
Another recent study~\cite{vuillod2024handling} proposes a Globally Convex Moving Mesh Algorithm-based knot vector optimization for Non-Uniform Rational B-Splines (NURBS) that uses an objective function combining data fitting and smoothing terms to reduce noise and prevent overfitting.
An earlier study~\cite{dertimanis2018data} determines optimal knot vectors and expansion coefficients by minimizing an objective function based on prediction error and smoothness using evolutionary strategies and interior-point algorithms.
A separate study work~\cite{dixler2021uncertainty} demonstrates high accuracy in numerical examples by employing sequential quadratic programming (SQP) to determine optimal knot vectors.
These methods, however, often require iterative optimization procedures that repeat evaluations of objective functions, which can be computationally demanding, particularly when integrated into design optimization workflows or applied to high-dimensional UQ problems. Another study~\cite{lee2022robust} uses SDD for solving robust design optimization, but does not consider optimal knot vectors. Therefore, we need to develop a more computationally efficient and accurate method for determining optimal knot vectors.

This work introduces a novel method for determining optimal knots in SDD for forward UQ in dynamical systems without relying on optimization procedures or requiring additional training data. 
In the method, we interpolate a response profile using a sample dataset that is used for training SDD. We then define reference regions based on the specified number of internal knots. 
Within each reference region, we identify and set the point where the gradient of the interpolated profile reaches its maximum as an internal knot. 
We then use these internal knots to construct orthonormal spline basis functions for SDD via a whitening transformation. 
To estimate the expansion coefficients of SDD, we use both standard least squares (SLS) and least absolute shrinkage and selection operator (LASSO) regression, with the latter being particularly effective for high-dimensional problems where SLS may suffer from overfitting or ill-conditioning. 

This paper is structured as follows. Section~\ref{sec:2} provides the theoretical background, including the definitions of input and output random variables and an introduction to the stochastic dynamic problem. In addition, the summary of the SDD principle, second-moment properties, and failure probability. Section~\ref{sec:3} proposes the determination of optimal knot vector and an SDD with the proposed knot vector. Also, convergence and stochastic properties are presented.  Section~\ref{sec:4} reports numerical examples. In Section~\ref{sec:5}, we offer conclusions and an outlook for future work.

\section{Theoretical background} \label{sec:2} 

\subsection{Preliminaries}\label{sec:2.0}
Let $\nat$, $\nat_{0}$, $\real$, and $\real_{0}^{+}$ denote the sets of positive integers, non-negative integers, real numbers, and non-negative real numbers, respectively. For a positive integer $N\in\nat$, let $\mathbb{A}^N \subseteq \real^N$ denote a bounded or unbounded sub-domain of $\real^N$. 
%
\subsection{Input random variables} \label{sec:2.1} 
Let $(\Omega,\cF,\mathbb{P})$ be an abstract probability space, with a sample space $\Omega$, a $\sigma$-algebra $\cF$ on $\Omega$, and a probability measure $\mathbb{P}:\cF\to[0,1]$. Consider an $N$-dimensional random vector $\bX:=(X_{1},\ldots,X_{N})^\intercal:\Omega \rightarrow \mathbb{A}^N$, that models the uncertainties in a stochastic problem. We refer to $\bX$ as the random input vector or the input random variables. 
Denote by $F_{\bX}({\bx}):=\mathbb{P}\big[\cap_{i=1}^{N}\{ X_i \le x_i \}\big]$ the joint cumulative distribution function (CDF) of $\bX$, admitting the joint probability density function (PDF) $f_{\bX}({\bx}):={\partial^N F_{\bX}({\bx})}/{\partial x_1 \cdots \partial x_N}$. For $(\Omega,\cF,\mathbb{P})$, the image probability space is $(\mathbb{A}^N,\cB^{N},f_{\bX}(\bx)\rd\bx)$, where $\mathbb{A}^N$ is the image of $\Omega$ under the mapping $\bX:\Omega \to \mathbb{A}^N$ and $\cB^N:=\cB(\mathbb{A}^N)$ is the Borel $\sigma$-algebra on $\mathbb{A}^N\subset \mathbb{R}^N$. \medskip

\subsection{Output random variables}  \label{sec:2.2}
Given an input random vector $\bX$
with a known probability measure $f_{\bX}({\bx})\mathrm{d}\mathbf{x}$ on $\mathbb{A}^N \subseteq \real^N$, denote by $y(\bX)$ a real-valued, square-integrable  transformation on $(\Omega, \cF)$.  Here, $y:\mathbb{A}^N \to \real$ represents a quantity of interest (QoI) used to estimate statistical measures (e.g., mean and variance) and failure probabilities. In this work, we assume that $y$  has a finite variance.

\subsection{Stochastic dynamic problem} \label{sec:2.3} 

Consider a linear, $M$-degree-of-freedom dynamical system with a stochastic mass matrix $\mathbf{M}(\mathbf{X}) \in \mathbb{R}^{M \times M}$, a stochastic damping matrix $\mathbf{C}(\mathbf{X}) \in \mathbb{R}^{M \times M}$, and a stochastic stiffness matrix $\mathbf{K}(\mathbf{X}) \in \mathbb{R}^{M \times M}$. The mass matrix $\mathbf{M}(\mathbf{X})$ is real, symmetric, and positive-definite, while the stiffness and damping matrices, $\mathbf{K}(\mathbf{X})$ and $\mathbf{C}(\mathbf{X})$, are real, symmetric, and positive semi-definite. The damping matrix $\mathbf{C}(\mathbf{X})$ can be proportional or non-proportional; if proportional, it is expressed as a linear combination of the mass and stiffness matrices. 
When subjected to an $M$-dimensional deterministic force vector $\mathbf{f}(t) \in \mathbb{R}^{M}$, the system's equation of motion in the time domain $t \in [0, T],~T\in \mathbb{R}^+$, is
\begin{align}
    \mathbf{M}(\mathbf{X}) \ddot{\mathbf{z}}(t; \mathbf{X}) 
+ \mathbf{C}(\mathbf{X}) \dot{\mathbf{z}}(t; \mathbf{X}) 
+ \mathbf{K}(\mathbf{X}) \mathbf{z}(t; \mathbf{X}) 
= \mathbf{f}(t).
\label{eq:1}
\end{align}
Here $\mathbf{z}(t; \mathbf{X})$ denotes the $M$-dimensional displacement vector; 
$\dot{\mathbf{z}}(t; \mathbf{X})$ represents the $M$-dimensional velocity vector; 
and $\ddot{\mathbf{z}}(t; \mathbf{X})$ signifies the $M$-dimensional acceleration vector. 
We focus on solving~\eqref{eq:1} for harmonic response and modal analysis, two prominent challenges in structural dynamics UQ problems. 

\begin{itemize}
   \item \textbf{Harmonic response:}  
        Consider a deterministic harmonic excitation described by the complex-valued force vector
        $\mathbf{f}(t) = \mathbf{F}(\omega)\exp(i\omega t)$,
        where $i = \sqrt{-1}$, $\omega \in [\omega_l, \omega_r] \subset \mathbb{R}^+$, $0 \leq \omega_l < \omega_r < \infty$, is the excitation angular frequency, and $\mathbf{F}(\omega) \in \mathbb{R}^M$ represents the real-valued force amplitude vector. The governing equation of motion in the frequency domain is    
        \begin{align}
            \left[ -\omega^2\mathbf{M}(\mathbf{X}) + i\omega\mathbf{C}(\mathbf{X}) + \mathbf{K}(\mathbf{X}) \right]\mathbf{Z}(\omega; \mathbf{X}) = \mathbf{F}(\omega),
        \label{eq:4}
        \end{align}
        where $\mathbf{Z}(\omega; \mathbf{X}) \in \mathbb{C}^M$ is the complex-valued displacement amplitude vector.

        In UQ analysis, the goal is to propagate the input randomness of $\mathbf{X}$ to the frequency response functions (FRFs), enabling the probabilistic characterization of $\mathbf{Z}(\omega; \mathbf{X})$.
   
   \item \textbf{Modal analysis:}
   For the equation of motion~\eqref{eq:1}, consider a general nonlinear eigenvalue problem of the form 
    \begin{align}
        f(\lambda(\mathbf{X}); \mathbf{M}(\mathbf{X}), \mathbf{C}(\mathbf{X}), \mathbf{K}(\mathbf{X}))\boldsymbol{\phi}(\mathbf{X}) = \mathbf{0},
        \label{eq:5}
    \end{align}
     where yields a random eigenvalue $\lambda(\mathbf{X}) \in \mathbb{R}$ or $\mathbb{C}$ and a random eigenvector $\boldsymbol{\phi}(\mathbf{X}) \in \mathbb{R}^M$ or $\mathbb{C}^M$. The specific form of the function $f(\cdot)$ depends on the application, and thus lead to a variety of linear or nonlinear eigenvalue problems. This work focuses exclusively on linear and quadratic eigenvalue problems. The eigensolutions can be obtained by 
    \begin{align}
        \det[f(\lambda(\mathbf{X}); \mathbf{M}(\mathbf{X}), \mathbf{C}(\mathbf{X}), \mathbf{K}(\mathbf{X}))] = 0.
        \label{eq:6}
    \end{align}
    The primary aim of solving a random eigenvalue problem is to quantify the probabilistic characteristics of eigenpairs
    $\{\lambda^{(i)}(\mathbf{X}), \boldsymbol{\phi}^{(i)}(\mathbf{X})\}, \quad i = 1, \dots, M,$ based on the known probability law of the input random vector $\mathbf{X}$.     

\end{itemize}

\subsection{Spline dimensional decomposition (SDD)} \label{sec:2.5} 

For the coordinate direction $k = 1,...,N$, 
$p_k\in \mathbb N_0$ is the order of polynomial, $n_k\ge (p_k +1)$ is the total number of basis function, the knot vector $\xi_k$ follows
\begin{align}
    \xi_k = \lbrace\xi_{k,i_k} \rbrace_{i_k=1}^{n_k+p_k+1} = \lbrace a_k = \xi_{k,1},\xi_{k,2},...,\xi_{k,n_k+p_k+1} =b_k \rbrace.
    \label{eq:8} 
\end{align}
Here, $\xi_{k,i_k}$ is the $i_k$th knot with $i_k=1,2,\ldots,n_p+p_k+1$ for the interval $[a_k,b_k]\subset \mathbb{R}$. 
The B-spline basis functions $B_{i_k,p_k,\xi_k}^k(x_k)$, $i_k=1,\ldots,n_k$, are constructed from the knot vector, as detailed in Appendix~\ref{sec:appx1}.  
   
These B-splines are not orthonormal with respect to the probability measure $f_{X_{k}}(x_k)\mathrm{d}x_k$ of $X_k$. A linear transformation introduced in~\cite{rahman2020spline} yields an $n_k$-dimensional vector of the orthonormal B-spline $\boldsymbol {\psi}_k(X_k)$ via a three step algorithm:

\begin{enumerate}
    \item[(1)] Starting with a set of B-spline of degree $p_k$, define an auxiliary set as an $n_k$-dimensional vector  \begin{align}
    \boldsymbol{P}_k(X_k) := (1, \, B_{2,p_k,\xi_k}^{k} (X_k), \dots, \, B_{n_k,p_k,\xi_k}^{k} (X_k))^\top,
    \label{eq:9} 
    \end{align} 
    where the first element is \emph{one} and the remaining elements can be arranged in any order without loss of generality.  
    %
    \item[(2)] 
    Construct an \( n_k \times n_k \) spline moment matrix
    \begin{align}
    \boldsymbol{G}_k := \mathbb{E} \left[ \boldsymbol{P}_k(X_k) \boldsymbol{P}_k^\top(X_k) \right].
    \label{eq:10} 
    \end{align}
    Here, $\mathbf{G}_k$ is symmetric and positive-definite, allowing for Cholesky factorization 
    $\boldsymbol{W}_k^\top \boldsymbol{W}_k = \boldsymbol{G}_k ^{-1}$,
    where $\boldsymbol{W}_k$ is a non-singular whitening matrix. 
    \item[(3)] 
    Construct an \( n_k \)-dimensional vector of orthonormal B-spline $\boldsymbol {\psi}_k(X_k)$ via a whitening transformation, i.e.,  
    \begin{align}
    \boldsymbol {\psi}_k(X_k) = \boldsymbol{W}_k\boldsymbol{P}_k(X_k).
    \label{eq:12}
    \end{align}   
    \end{enumerate} 

For a subset $\emptyset \neq u = \{k_1, \dots, k_{|u|}\} \subseteq \{1, \dots, N\}$ with cardinality $1 \le k_{|u|} \le N.$ Let $X_u := \left(X_{k_1}, \dots, X_{k_{|u|}} \right)^\top$ be a subvector of $\boldsymbol{X}$ on $\left (\Omega^u, \mathcal{F}^u, \mathbb{P}^u \right)$, where $\Omega^u$ is the sample space of $X_u$, $\mathcal{F}^u$ is $\sigma-$algebra on $\Omega^u$, and $\mathbb{P}^u$ is a probability measure. As $\boldsymbol{X}$ comprises independent random variables, the joint PDF of $\mathbf{x}_u$ is
    $f_{\mathbf{X}_u}(\mathbf{x}_u) = \prod_{l=1}^{|u|} f_{X_{k_l}}(x_{k_l}),~
    \mathbf{x}_u := (x_{k_1}, \dots, x_{k_{|u|}})^\top$.
    
We define three multi-indices in all coordinate directions $|u|$:  knot indices $\boldsymbol{i}_u:=(i_{k_1}, \dots, i_{k_{|u|}})\in \mathbb{N}^{|u|}$ numbers of basis functions $\boldsymbol{n}_u := (n_{k_1}, \dots, n_{k_{|u|}}) \in \mathbb{N}^{|u|}$, and spline degrees $\boldsymbol{p}_u := (p_{k_1}, \dots, p_{k_{|u|}}) \in \mathbb{N}_0^{|u|}$.
Denote by $\boldsymbol{\Xi}_u := \lbrace \xi_{k_1}, \dots, \xi_{k_{|u|}} \rbrace $ a family of all $|u|$ knot sequences. Associated with $ \mathbf{i}_u$, define an index set 
    \begin{align}
    \mathcal{I}_{u, \mathbf{n}_u} := \left\{ \mathbf{i}_u = (i_{k_1}, \dots, i_{k_{|u|}}) : 1 \leq i_{k_l} \leq n_{k_l}, \, l = 1, \dots, |u| \right\} \subset \mathbb{N}^{|u|}
    \label{eq:14}
    \end{align}
    with cardinality $
        |\mathcal{I}_{u, \mathbf{n}_u}| = \prod_{k \in u} n_k$.

For the coordinate direction $k_l$, let
$I_{k_l} = r_{k_l} -1$ denote the number of subintervals corresponding to the knot vector $\mathbf{\xi}_{k_l}$ with $r_{k_l}$ distinct knots. The knot vectors $\xi_{k_1}, \dots, \xi_{k_|u|}$ define a partition of the $|u|$-dimensional rectangle $\mathbb{A}^u := \times_{k \in u} a_k, b_k$, decomposing it into smaller rectangles 
    \begin{align}
        \left\{ 
        \mathbf{x}_u = (x_{k_1}, \dots, x_{k_{|u|}}) : 
        \xi_{k_l, j_{k_l}} \leq x_{k_l} < \xi_{k_l, j_{k_l}+1}, \, 
        l = 1, \dots, |u| 
        \right\}, \, 
        j_{k_l} = 1, \dots, I_{k_l},
    \end{align}
where $\xi_{k_l, j_{k_l}}$ is the $j_{k_l}$th distinct knot in the coordinate direction $k_l$. The partitioning of $\mathbb{A}^u$ into these rectangular elements defines a mesh.
    
As a result, the multivariate orthonormal B-splines in $\mathbf{x}_u = (x_{k_1}, \dots, x_{k_{|u|}})$ consistent with the probability measure $f_{\mathbf{X}_u}(\mathbf{x}_u) \, d\mathbf{x}_u$ are
    \begin{align}
        \Psi_{\mathbf{i}_u, \mathbf{p}_u, \mathbf{\Xi}_u}^u (\mathbf{x}_u) 
        = \prod_{l=1}^{|u|} \psi_{i_{k_l}, p_{k_l}, \xi_{k_l}}^{k_l} (x_{k_l}), 
        \quad \mathbf{i}_u = (i_{k_1}, \dots, i_{k_{|u|}}) \in \bar{\mathcal{I}}_{u, \mathbf{n}_u},
    \label{eq:20}
    \end{align}
    where $\bar{\mathcal{I}}_{u, \mathbf{n}_u} := 
        \left\{ 
        \mathbf{i}_u = (i_{k_1}, \dots, i_{k_{|u|}}) : 
        2 \leq i_{k_l} \leq n_{k_l}, \, l = 1, \dots, |u| 
        \right\} \subset \left(\mathbb{N} \setminus \{1\}\right)^{|u|}$ 
    is a reduced index set, which has cardinality
$\left| \bar{\mathcal{I}}_{u, \mathbf{n}_u} \right| := \prod_{k \in u} (n_k - 1)$.

The first and second order moments of multivariate orthonormal B-splines are
\begin{equation}\label{eq:12} 
\begin{aligned}
    &\mathbb{E} \left[ \Psi_{\mathbf{i}_u, \mathbf{p}_u, \mathbf{\Xi}_u}^u (\mathbf{X}_u) \right] = 0 
    \quad \text{and}& &
    \mathbb{E} \left[ \Psi_{\mathbf{i}_u, \mathbf{p}_u, \mathbf{\Xi}_u}^u (\mathbf{X}_u) 
     \Psi_{\mathbf{j}_v, \mathbf{p}_v, \mathbf{\Xi}_v}^v (\mathbf{X}_v) \right] =
    \begin{cases}
        1, & u = v \text{ and } \mathbf{i}_u = \mathbf{j}_v, \\
        0, & \text{otherwise},
    \end{cases}
\end{aligned}
\end{equation} 
respectively.        
We can represent any random variable $y(\mathbf{X})$ using a hierarchical Fourier-like expansion in multivariate orthonormal B-splines. In practices, this expansion involves at most $1 \le S \le N$ variables, leading to an $S$-variate SDD approximation~\cite{rahman2022spline}, i.e., 
    \begin{align}
        y_{S, \mathbf{p}, \mathbf{\Xi}}(\mathbf{X}) := y_{\emptyset} + 
        \sum_{\substack{\emptyset \neq u \subseteq \{1, \dots, N\} \\ 1 \leq |u| \leq S}} 
        \sum_{\mathbf{i}_u \in \bar{\mathcal{I}}_{u, \mathbf{n}_u}} 
        c_{\mathbf{i}_u, \mathbf{p}_u, \mathbf{\Xi}_u}^u 
        \Psi_{\mathbf{i}_u, \mathbf{p}_u, \mathbf{\Xi}_u}^u (\mathbf{X}_u)
    \label{eq:25}
    \end{align}

    of $y(\boldsymbol{X}) $, comprising basis functions are \begin{align} 
        L_{S, \mathbf{p}, \mathbf{\Xi}} = 1 + \sum_{\substack{\emptyset \neq u \subseteq \{1, \dots, N\} \\ 1 \leq |u| \leq S}} \prod_{k \in u} (n_k - 1) \leq \prod_{k=1}^N n_k.
        \label{num_SDD}
        \end{align} 
        Readers are encouraged to refer to the cited work~\cite{rahman2022spline} for further details.

\subsection{Second-moment properties and failure probability via SDD}\label{sec:2.6} 

Using the first and second order moments~\eqref{eq:12} of multivariate orthonormal B-splines, we derive analytical expressions for the mean and variance as  
    \begin{align}
    \mathbb{E} \left[ y_{S, \mathbf{p}, \mathbf{\Xi}}(\mathbf{X}) \right] = y_{\emptyset} = \mathbb{E} \left[ y(\mathbf{X}) \right]\quad\text{and}\quad     \mathrm{var} \left[ y_{S, \mathbf{p}, \mathbf{\Xi}}(\mathbf{X}) \right] = 
    \sum_{\substack{\emptyset \neq u \subseteq \{1, \dots, N\} \\ 1 \leq |u| \leq S}}
    \sum_{\mathbf{i}_u \in \bar{\mathcal{I}}_{u, \mathbf{n}_u}}
    \left( c^{u}_{\mathbf{i}_u, \mathbf{p}_u, \mathbf{\Xi}_u} \right)^2 
    \label{eq:31}
    \end{align}
    
    of $y_{S, \mathbf{p}, \mathbf{\Xi}}(\mathbf{X})$, respectively. 
   
        Let $\Omega_{F,S, \mathbf{p}, \mathbf{\Xi}} := \{\mathbf{x} : y_{S, \mathbf{p}, \mathbf{\Xi}}(\mathbf{x}) < 0\},$ be an approximate failure set as a result of an SDD approximation $y_{S, \mathbf{p}, \mathbf{\Xi}}(\mathbf{X})$ of $y(\mathbf{X})$. Then, we estimate the failure probability $P_F$ using MCS of the SDD approximation, i.e.,
        \begin{align}
            P_{F,S, \mathbf{p}, \mathbf{\Xi}} = \mathbb{E}\left[ I_{\Omega_{F,S, \mathbf{p}, \mathbf{\Xi}}}(\mathbf{X}) \right] = \lim_{L' \to \infty} \frac{1}{L'} \sum_{l=1}^{L'} I_{\Omega_{F,S, \mathbf{p}, \mathbf{\Xi}}} \big( \mathbf{x}^{(l)} \big).
            \label{eq:33}
        \end{align}
        Here, $L'$ is the sample size of the SDD approximation, $\mathbf{x}^{(l)}$ is the $l$th realization of $\mathbf{X}$, and $I_{\Omega_{F,SDD}}(\mathbf{x})$ is another indicator function, which is equal to $one$ when $\mathbf{x} \in \Omega_{F,SDD}$ and $zero$ otherwise.

\begin{problem}\label{prob}
Consider the accuracy of SDD approximation, which heavily depends on the selection of the knot vector. Given a stochastic dynamical system and a computationally intensive model, the objective is to efficiently determine an optimal knot vector to enhance the accuracy of the SDD approximation for stochastic dynamic analysis.
\end{problem} 

\section{Interpolation-based optimal spline dimensional decomposition}\label{sec:3}

\subsection{Non-uniform vs. uniform knot vectors in SDD} \label{sec:3.1}
\begin{figure} [!ht]
    \begin{center}
    \includegraphics[angle=0,scale=0.4,clip]{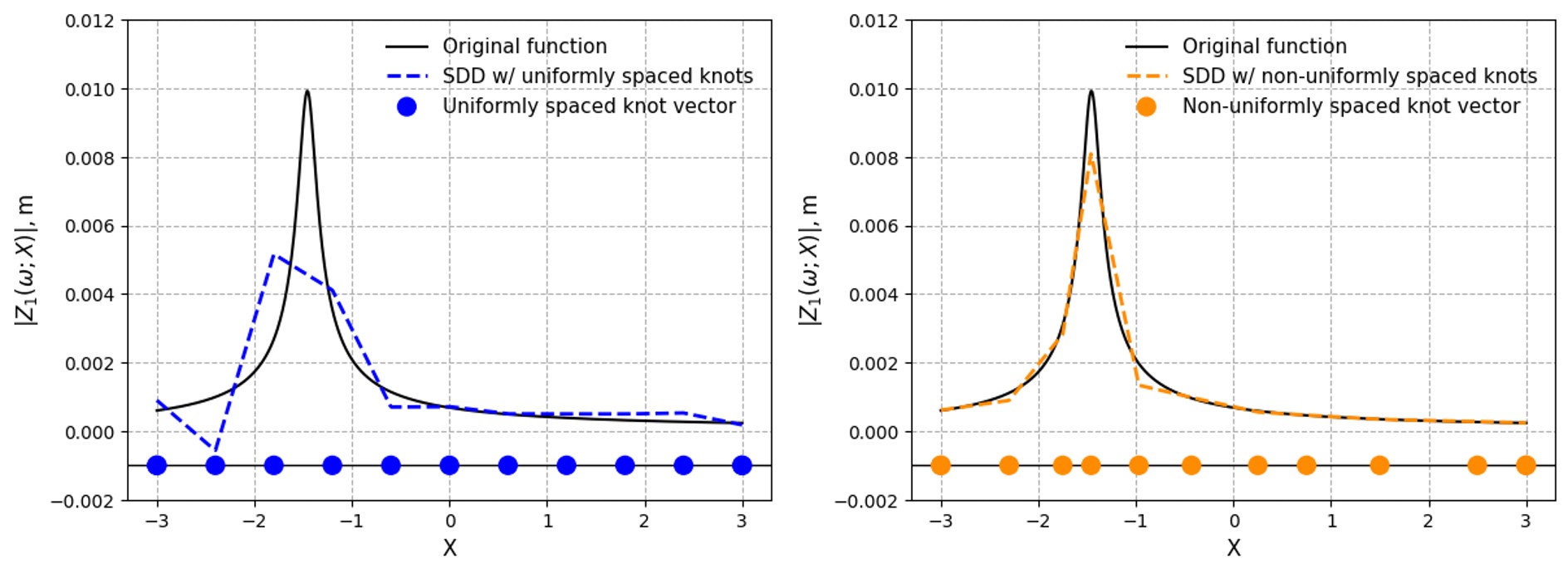}
    \end{center}
    \caption {Comparison of SDD approximations ($p=1, I=10$) using two different knot vector:
    (left) a uniformly spaced knot $\xi = \{-3, -3,-2.4, -1.8, -1.2, -0.6, 0, 0.6, 1.2, 1.8, 2.4, 3, 3\}$,
    and (right) a non-uniformly spaced knot $\xi = \{-3, -3, -2.3, -1.75, -1.46, -0.97, -0.43, 0.25, 0.75, 1.5, 2.5, 3, 3\}$. The approximated responses are compared with the original function to evaluate the impact of knot placement on the accuracy of the SDD}
    \label{fig1}
    \end{figure}

   The placement of internal knots plays a critical role in determining the nonsmoothness and local profile of each B-spline basis function $B_{i_k,p_k,\xi_k}(x_k)$, which is nonzero within the interval $[\xi_{k,i_k}, \xi_{k,i_k+p_k+1})$. This local support property implies that the overall approximation space is composed of piecewise polynomials defined over these intervals. Consequently, the accuracy of the SDD approximation depends on how effectively the B-spline basis functions capture the local behavior of the target response. Inappropriate knot placement may yield basis functions that are unable to accurately represent nonlinear or localized variations in the response~\cite{dixler2021uncertainty}. Figure~\ref{fig1} illustrates the harmonic response of a univariate input $X\in[-3,3]$ for two cases of knot vectors: uniformly spaced and non-uniformly spaced. 
   For the non-uniform case, a knot is strategically placed near the response peak at $X=-1.46$, which corresponds to a region of high gradient variation. SDD approximations are generated with both knot configurations ($p=1, I=10$) and compared graphically against against the original function. 
   The SDD with the non-uniform knot vector captures the non-smooth region more effectively than its uniform counterpart.  
   %
   While uniform knot vectors are often chosen for simplicity, they may not adequately represent localized features in highly nonlinear responses. 
   As B-spline basis functions are directly controlled by knot placement, suboptimal knot positioning can reduce accuracy locally and increase overall error. Consequently, determining the optimal knot vector based on the response characteristics is crucial for improving the SDD model's accuracy.

    \subsection{Interpolation method for an optimal knot vector}\label{sec:3.2} 

\begin{figure} [htbp]
    \begin{center}
    \includegraphics[angle=0,scale=0.4,clip]{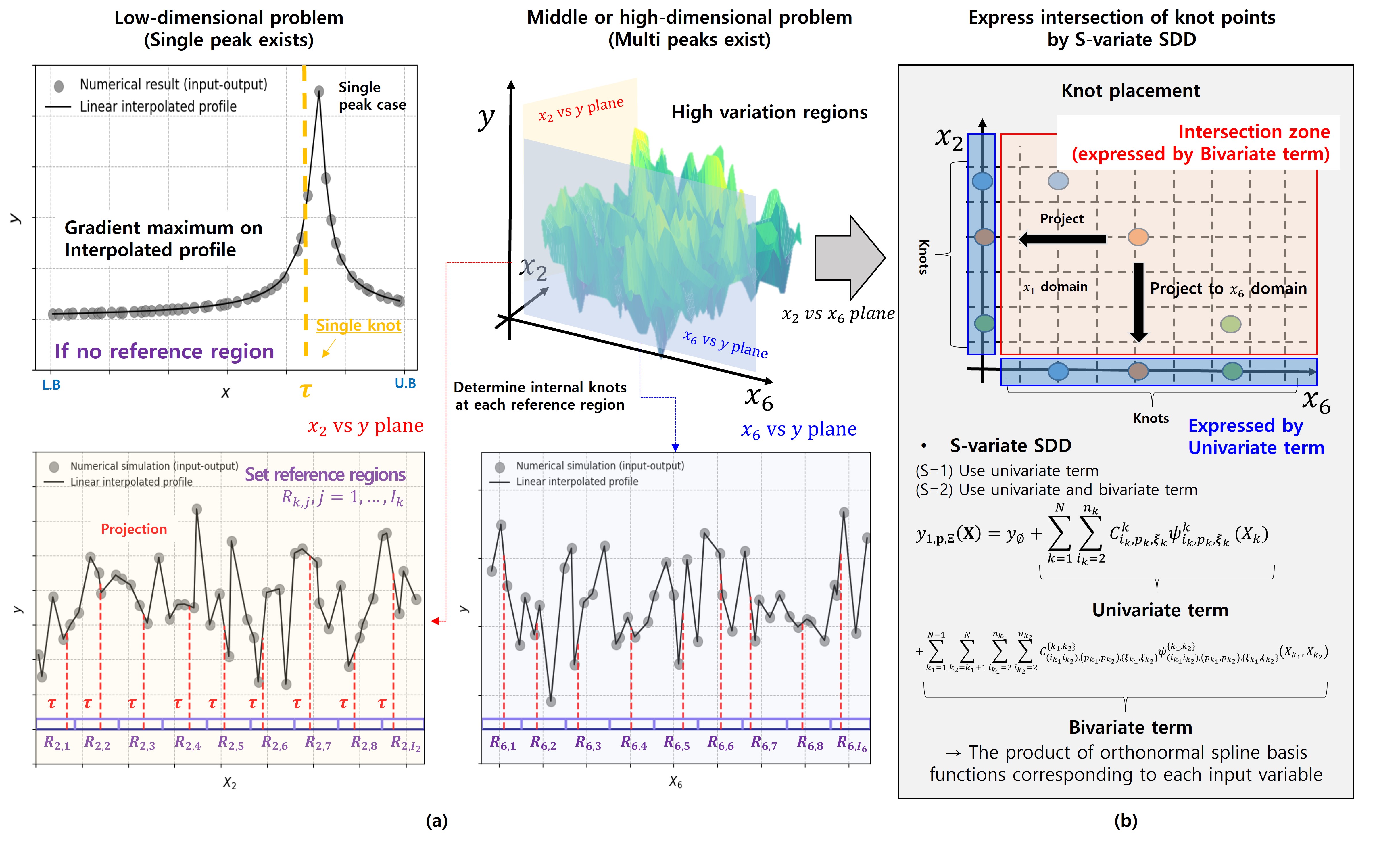}
    \end{center}
    \caption {(a) In low-dimensional problems, the response typically exhibits a single dominant peak. However, in medium- or high- dimensional problems, multiple peaks are common. To effectively capture such localized nonlinear behaviors, the input domain is uniformly partitioned into reference regions, each responsible for identifying significant local variations; (b) When $S=1$, the approximation consists of univariate terms constructed from orthonormal B-spline basis functions associated with individual knot locations. When $S=2$, bivariate terms are additionally included to represent interaction zones, where the basis is formed by the product of two orthonormal spline basis functions. These structured terms enable accurate representation of variable interactions in high-dimensional problems.}
    \label{fig2}
    \end{figure}

   In the proposed method, reference regions are introduced to address high-dimensional problems where the response shows complex oscillatory behavior. As illustrated in Figure~\ref{fig2}, such behavior includes multiple local peaks. This makes it difficult to capture significant variation with a single representative knot. To overcome this limitation, multiple internal knots should be identified rather than relying on a single representative point. Instead of searching across the entire input domain at once, it is more effective to divide the domain into local regions and identify the maximum gradient point within each region. Based on this idea, the input space is uniformly partitioned into reference regions $R_{k,j}$. This uniform division ensures that the entire domain is equally considered without additional complexity. The number of reference regions is set equal to the number of internal knots, allowing each internal knot to capture the most significant variation within its region by gradient method.

   Algorithm~\ref{al1} outlines the implementation steps of the proposed interpolation-based knot selection procedure. We obtain input-output data samples $\{\mathbf{x}^{(l)},y^{(l)}\}_{l=1}^L$, where $y^{(l)}=y(\mathbf{x}^{(l)})$, from either experimental measurements or numerical simulations. 
   For the $k$th coordinate, we rearrange the dataset so that the input values are ordered as $x_k^{(l_1)}< \cdots < x_k^{(l_L)}$, with corresponding output values $y^{(l_1)},\ldots,y^{(l_L)}$, where $\{l_1,\ldots,l_L\}$ is a permutation of the index set $\{1,\ldots,L\}$.
   The output profile is estimated using linear interpolation of the sampled input-output pairs, obtaining a linear function $f(x_k)$, as detailed in Algorithm~\ref{al1}. The input domain is then partitioned into $I_k$ reference regions $R_{k,j}$, i.e,
   \begin{align*}
   R_{k,j} \in 
   \left[
      a_k + \frac{(j-1)(b_k - a_k)}{I_k}, \ 
      a_k + \frac{j(b_k - a_k)}{I_k}
   \right], \quad j = 1, \dots, I_k.
   \end{align*}
   Within each region, the maximum gradient point of the interpolated response is identified and designated as the internal knot $\tau_{k,j}$, i.e, 
   \begin{align}\label{int_knot}
    \tau_{k,j} 
    = \argmax_{x_k^{(l_i)}\in R_{k,j}} 
    \left| \frac{f(x_k^{(l_i)}) - f(x_k^{(l_{i-1})})}{x_k^{(l_i)} - x_k^{(l_{i-1})}} \right|,
    \quad i=2,\ldots,L,~j = 1, \dots, I_k,
    \end{align}
where $x_k^{(l_{i-1})}< x_k^{(l_i)}$ and $l_i\in\{1,\ldots,L\}$. This knot selection procedure does not require any additional computational cost, as it operates directly on existing samples $\{\mathbf{x}^{(l)},y^{(l)}\}_{l=1}^L$ used for estimating the SDD coefficients (see Section~\ref{sec:3.3.3}). The procedure also avoids additional optimization steps, thereby enabling efficient identification of nonlinear behavior within each subdomain.

   
   The identified internal knots $\tau_{k,j}$, $j=1,\ldots,I_k$, are finally combined with the boundary values $a_k$ and $b_k$ to construct the optimal knot vector $\xi_k^*$. The resulting vector satisfies all knot properties defined in Section~\ref{sec:2.5} and contributes to improving the accuracy of the SDD approximation. 
    
   \begin{algorithm}
    \caption{Determine internal knots $\tau_{k,j}$ and optimal knot vector $\xi_k^*$.}
    \label{al1}
    \begin{algorithmic}[1] 
    \Require Set a knot vector $\boldsymbol{\Xi}_u = (\xi_{k_1}, \dots, \xi_{k_{|u|}})^\intercal$, uniformly spaced in this work. Samples $\bx^{(l)}=(x_1^{(l)},\ldots,x_N^{(l)})^{\intercal}$, $l=1,\ldots, L$, via MCS, quasi MCS, or Latin hypercube sampling with corresponding probabilities $p^{(l)}=f_{\bX}(\bx^{(l)}){\rm d} \bx^{(l)}$; Generate the input-output samples $\{\bx^{(l)},y(\bx^{(l)})\}_{l=1}^{L}$. 
    
    \Ensure  Optimal knot vector $\xi_k^*$.
    \State For each $k$th coordinate, rearrange the input-output samples $\{\bx^{(l)},y^{(l)}\}_{l=1}^{L}$ so that $x_k^{(l_1)}<\cdots<x_k^{(l_L)}$, with the corresponding outputs $y^{(l_1)},\ldots,y^{(l_L)}$.

    \State Estimate output profile by a linear interpolation method, $i =  2,\dots,L$, $k=1,\dots,N$,
    \[
        \quad f(x_k) = y^{(l_{i-1})} + \frac {y^{(l_i)}-y^{(l_{i-1})}}{x^{(l_i)}_k-x_k^{(l_{i-1})}} (x_k-x_k^{(l_{i-1})}),
        \quad x_k^{(l_{i-1})} \leq x_k \leq x_k^{(l_i)}.
    \]

    \State Determine the number of subintervals $I_k$.
    \State Set reference regions $R_{k,j}$ as

    \begin{align*}
   R_{k,j} \in 
   \left[
      a_k + \frac{(j-1)(b_k - a_k)}{I_k}, \ 
      a_k + \frac{j(b_k - a_k)}{I_k}
   \right], \quad j = 1, \dots, I_k.
   \end{align*}

    \State Identify internal knots $\tau_{k,j}$ with maximum gradient of the profile on $R_{k,j}$
   
    \begin{align*}
    \tau_{k,j} 
    = \argmax_{x^{(l_i)}\in R_{k,j}} 
    \left| \frac{f(x^{(l_i)}) - f(x^{(l_{i-1})})}{x^{(l_i)} - x^{(l_{i-1})}} \right|,
    \quad i=2,\ldots,L,~j = 1, \dots, I_k.
    \end{align*}

    \State Construct the optimal knot vector $\xi_k^*$,
    
    \[\xi_k^* = \{a_k = \xi_{k,1}^*, \xi_{k,2}^*, \xi_{k,3}^*,\dots,\xi_{k,n+p-1}^*,\xi_{k,n+p}^*,\xi_{k,n+p+1}^* = b_k\},
    \]    

    \[
        \text{where }
        \tau_{k,1}, \tau_{k,2},\dots,\tau_{k,I_k-1},\tau_{k,I_k} = \xi_{k,2}^*,\xi_{k,3}^*,\dots,\xi_{k,n+p-1}^*,\xi_{k,n+p}^*.
    \]
    
    \end{algorithmic}
    \end{algorithm}
   
\subsection{Spline dimensional decomposition with optimal knot vector}\label{sec:3.3}  
We employ the optimal knot vector, as determined by Algorithm~\ref{al1}, to construct the S-variate SDD approximation. This method enables more accurate solutions for the high-dimensional dynamic problems outlined in Problem~\ref{prob}.

\subsubsection{Univariate orthonormal B-splines} \label{sec:3.3.1} 
We derive the univariate orthonormal B-spline basis functions $\psi_{i_k,p_k,\xi_k^*}^{k}(x_k)$ in $X_k$, $k=1,\ldots,N$, via a whitening transformation of the standard B-splines defined on the optimal knot vector $\xi_k^*$ as introduced in Section~\ref{sec:3.2}. This transformation ensures orthonormality with respect to the probability measure $f_{X_k}(x_k)\mathrm{d}x_k$ associated with the input random variable $X_k$, following the three-step procedure described in Section~\ref{sec:2.5}. Under this measure, the resulting basis functions satisfy the second moment properties in \eqref{eq:31}.




\subsubsection{Multivariate orthonormal B-splines} \label{sec:3.3.2} 

We construct the multivariate orthonormal B-splines in $\mathbf{X}_u = (X_{k_1}, \dots, X_{k_{|u|}})$ consistent with the probability measure $f_{\mathbf{X}_u}(\mathbf{x}_u) \, d\mathbf{x}_u$ as 
    \begin{align}
        \Psi_{\mathbf{i}_u, \mathbf{p}_u, \mathbf{\Xi}_u^*}^u (\mathbf{x}_u) 
        = \prod_{k \in u} \psi_{i_k, p_k, \xi_k^*}^k (x_k) 
        = \prod_{l=1}^{|u|} \psi_{i_{k_l}, p_{k_l}, \xi_{k_l}^*}^{k_l} (x_{k_l}), 
        \quad \mathbf{i}_u = (i_{k_1}, \dots, i_{k_{|u|}}) \in \bar{\mathcal{I}}_{u, \mathbf{n}_u}.
    \label{eq:36}
    \end{align}
The first and second order moments of these B-splines $\Psi_{\mathbf{i}_u, \mathbf{p}_u, \mathbf{\Xi}_u^*}^u (\mathbf{x}_u)$ satisfy the orthonormality condition~\eqref{eq:12}.

We consider the set of orthonormalized B-splines $
\{\Psi_{\mathbf{i}_u, \mathbf{p}_u, \mathbf{\Xi}_u^*}^u (\mathbf{x}_u) :~ 1 \leq |u| \leq S,\ \mathbf{i}_u \in \bar{\mathcal{I}}_{u,\mathbf{n}_u}\}$,
which comprises $L_{S,\mathbf{p},\mathbf{\Xi}^*}$~\eqref{num_SDD} basis functions. We arrange this set as \begin{align}
\left\{\Psi_{\mathbf{i}_u, \mathbf{p}_u, \mathbf{\Xi}_u^*}^u (\mathbf{x}_u) :~ 1 \leq |u| \leq S,\ \mathbf{i}_u \in \bar{\mathcal{I}}_{u,\mathbf{n}_u} \right\}=\left\{\Psi_1(\mathbf{x};\mathbf{\Xi}^*) = 1,\Psi_2 (\mathbf{x};\mathbf{\Xi}^*),\ \ldots,\ \Psi_{L_{S,\mathbf{p},\mathbf{\Xi}^*}} (\mathbf{x};\mathbf{\Xi}^*) \right\},\label{rearrange_multiON}
\end{align} 
where $\Psi_i(\mathbf{x};\mathbf{\Xi}^*)$ denotes the $i$th basis function in the truncated SDD approximation.

\subsubsection{Calculation of expansion coefficients} \label{sec:3.3.3} 

For an output $y(\mathbf{X})$, the resulting S-variate SDD approximation with optimal knot vector $\mathbf{\Xi}^*$ is
   \begin{align}
        y_{S, \mathbf{p}, \mathbf{\Xi}^*}(\mathbf{X}) :=
        \sum_{i=1}^{L_{S,\mathbf{p},\mathbf{\Xi^*}}} 
        c_{i}(\mathbf{\Xi}^*) 
        \Psi_i (\mathbf{X};\mathbf{\Xi}^*),
        \label{eq:37}
    \end{align}
    where 
\begin{align*}
c_i(\mathbf{\Xi}^*)=\int_{\mathbb{A}^N}y(\mathbf{x})\Psi_{i}(\mathbf{x};\mathbf{\Xi}^*)f_{\mathbf{X}}(\mathbf{x})\mathrm{d}\mathbf{x}.
\end{align*} 

The expansion coefficients $c_i(\mathbf{\Xi}^*)$, $i=1,\ldots,L_{S,\mathbf{p},\mathbf{\Xi}^*}$ can be estimated using numerical integration methods, such as Gauss quadrature, or regression-based methods, including least absolute shrinkage and selection operator with cross-validation (LASSO-CV)~\cite{tibshirani1996regression} or standard least squares (SLS)~\cite{rahman2022spline}.
In this work, we focus on regression-based methods, which require fewer samples than numerical integration and are therefore more computationally efficient for UQ problems involving high-dimensional inputs~\cite{lee2020practical}. 
%

B-spline basis functions are nonzero only over specific intervals, so only a subset of them is active at any given input sample. This localized behavior makes the regression design matrix structurally sparse and increases the likelihood of sparsity in the model. LASSO-CV effectively handles spline-based models through an $\ell_1$-norm penalty. This penalty sets coefficients to zero for less significant basis functions while estimating coefficients for more influential ones. Thus, the SDD model avoids overfitting, reduces computational cost, and becomes more interpretable. In contrast, SLS is a widely adopted regression method, and its inclusion in this study demonstrates the applicability of the proposed framework even under general, non-sparse conditions.

 Consider an input-output dataset $\{\mathbf{x}^{(l)},y(\mathbf{x}^{(l)})\}_{l=1}^L$ of size \( L \in \mathbb{N} \). The output function \( y(\cdot) \) can be as simple as an explicitly defined mathematical function or as intricate as an implicitly described function obtained via computational simulation. In either case, the dataset, often referred to as the experimental design, can be generated by evaluating \( y(\mathbf{x}^{(l)}) \) at each input realization \( \mathbf{x}^{(l)} \). Various sampling methods, namely, standard MCS, quasi-MCS, and Latin hypercube sampling, can be employed to construct this dataset.
 \begin{itemize}  
   \item \textbf{LASSO-CV} 
    To estimate the expansion coefficients of the SDD approximation \( y_{S, \mathbf{p}, \mathbf{\Xi}^*}(\mathbf{X}) \), we use LASSO-CV. 
    The resulting coefficient vector \( \hat{\mathbf{c}}(\mathbf{\Xi}^*) \) provides a sparse and accurate representation of the response \( y(\mathbf{X}) \) while preventing overfitting.
    The expansion coefficients \( \hat{\mathbf{c}}(\mathbf{\Xi}^*)=(\hat{c}_1(\mathbf{\Xi}^*), \dots, \hat{c}_{L_{S, \mathbf{p}, \mathbf{\Xi}^*}}(\mathbf{\Xi}^*))^{\intercal} \) are obtained by 
\begin{align}
    \hat{\mathbf{c}}(\mathbf{\Xi}^*) = \underset{\mathbf{c} \in \mathbb{R}^{L_{S, \mathbf{p}, \mathbf{\Xi}^*}}}{\arg\min}
    \left\| \mathbf{b} - \bar{\mathbf{A}} \mathbf{c} \right\|_2^2 + \hat{\lambda} \left\| \mathbf{c} \right\|_1,
    \label{eq:42}
\end{align}
where \begin{align}
        \bar{\mathbf{A}} :=
        \begin{bmatrix}
        \Psi_1(\mathbf{x}^{(1)}; \mathbf{\Xi}^*) & \cdots & \Psi_{L_{s, \mathbf{p}, \mathbf{\Xi}^*}}(\mathbf{x}^{(1)}; \mathbf{\Xi}^*) \\
        \vdots & \ddots & \vdots \\
        \Psi_1(\mathbf{x}^{(L)}; \mathbf{\Xi}^*) & \cdots & \Psi_{L_{s, \mathbf{p}, \mathbf{\Xi}^*}}(\mathbf{x}^{(L)}; \mathbf{\Xi}^*)
        \end{bmatrix},
        \quad
        \mathbf{b} :=
        \begin{bmatrix}
        y(\mathbf{x}^{(1)}) \\
        \vdots \\
        y(\mathbf{x}^{(L)})
        \end{bmatrix}
        \label{eq:41}
    \end{align}
    are the design matrix and output vectors, respectively, composed of orthonormal basis function evaluations and their corresponding response values. We identify the optimal regularization parameter by minimizing the cross-validation error as  
    \begin{align}
    \hat{\lambda} = \underset{\lambda \in \{\lambda_1, \dots, \lambda_m\}}{\arg\min} \frac{1}{n} \sum_{k\in\{1:n\}} \sum_{j \in F_k} \left[ y(\mathbf{x}^{(j)}) - y_{S, \mathbf{p}, \mathbf{\Xi}^*}^{(-j)}(\mathbf{x}^{(j)}; \lambda) \right]^2.
    \label{eq:40}
    \end{align}
    Here $\{F_1,\ldots,F_n\}$ denotes a partition of the index set $\{1,\ldots,L\}$ into $n$ folds for $k$-fold cross-validation, and $y_{S,\mathbf{p},\mathbf{\Xi}^*}^{(-j)}$ is the prediction obtained by excluding fold $j$ from the training data.

    \item \textbf{SLS} 
    %
    %
    The coefficient vector is estimated by 
    \begin{align}
    \hat{\mathbf{c}}(\mathbf{\Xi}^*) = \underset{\mathbf{c} \in \mathbb{R}^{L_{S, \mathbf{p}, \mathbf{\Xi}^*}}}{\arg\min}
    \left\| \mathbf{b} - \bar{\mathbf{A}} \mathbf{c} \right\|_2^2,
    \label{eq:44}
\end{align}
where the matrices $\bar{\mathbf{A}}$ and $\mathbf{b}$ are defined in \eqref{eq:41}. The SLS solution is
    \begin{align}
    \hat{\mathbf{c}}(\mathbf{\Xi}^*) :=
    \begin{pmatrix}
    \hat{c}_1(\mathbf{\Xi}^*), \dots, \hat{c}_{L_{S, \mathbf{p}, \mathbf{\Xi}^*}}(\mathbf{\Xi}^*)
    \end{pmatrix}^\top
    = \left( \bar{\mathbf{A}}^\top \bar{\mathbf{A}} \right)^{-1} \bar{\mathbf{A}}^\top \mathbf{b}.
    \label{eq:45}
    \end{align}
 \end{itemize}

\subsubsection{Output statistics and other stochastic properties} \label{sec:3.2.4} 

\begin{itemize}    
   \item Mean and variance
    \hspace{1em} Applying the expectation operator on $y_{S, \mathbf{p}, \mathbf{\Xi}^*}(\mathbf{X})$ in~\eqref{eq:25} and recognizing~\eqref{eq:12}, the mean of SDD approximation is 
   
    \begin{align}
    \mathbb{E} \left[ y_{S, \mathbf{p}, \mathbf{\Xi}^*}(\mathbf{X}) \right] = y_{\emptyset} = \mathbb{E} \left[ y(\mathbf{X}) \right].
    \label{eq:46}
    \end{align}
    The SDD approximation always yields the exact mean, provided that the expansion coefficient $y_{\emptyset}$ is determined exactly. The variance of $y(\mathbf{x})$ is estimated by \medskip

    \begin{align}
    \mathrm{var} \left[ y_{S, \mathbf{p}, \mathbf{\Xi}^*}(\mathbf{X}) \right] = 
    \sum_{\substack{\emptyset \neq u \subseteq \{1, \dots, N\} \\ 1 \leq |u| \leq S}}
    \sum_{\mathbf{i}_u \in \bar{\mathcal{I}}_{u, \mathbf{n}_u}}
    \left( C^{u}_{\mathbf{i}_u, \mathbf{p}_u, \mathbf{\Xi}_u^*} \right)^2
    \leq \mathrm{var} \left[ y(\mathbf{X}). \right]
    \
    \label{eq:47}
    \end{align}
    
    The SDD approximation $y_{S, \mathbf{p}, \mathbf{\Xi}^*}(\mathbf{X})$ converges to the true response $y(\mathbf{X})$ in the mean-square sense as the mesh size $\mathbf{h} \to 0$ and the interaction order $S \to N$. This convergence is ensured by the use of hierarchical structure of the orthonormal B-spline basis, which enables efficient representation of the function space as the expansion grows.
    \item Failure probability
    Let $\Omega_{F,S, \mathbf{p}, \mathbf{\Xi}^*} := \left\{ \mathbf{x} \in \mathbb{A}^N : y_{S, \mathbf{p}, \mathbf{\Xi}^*}(\mathbf{x}) < 0 \right\}$ be the approximate failure set associated with the SDD approximation $y_{S, \mathbf{p}, \mathbf{\Xi}^*}(\mathbf{X})$ with proposed optimal knot vector $\mathbf{\Xi}^*$. Then, the failure probability is estimated as
    \begin{align}
    P_{F,S, \mathbf{p}, \mathbf{\Xi}^*} 
    = \mathbb{E} \left[ I_{\Omega_{F,S, \mathbf{p}, \mathbf{\Xi}^*}}(\mathbf{X}) \right] 
    \approx \frac{1}{L'} \sum_{l=1}^{L'} I_{\Omega_{F,S, \mathbf{p}, \mathbf{\Xi}^*}}(\mathbf{x}^{(l)}),
    \label{eq:FP_optknot}
    \end{align}
    
    where $L'$ is the number of resamples, $\mathbf{x}^{(l)}$ is the $l$th realization of $\mathbf{X}$, and $I_{\Omega_{F,S, \mathbf{p}, \mathbf{\Xi}^*}}(\mathbf{x})$ is the indicator function that equals $one$ when $\mathbf{x} \in \Omega_{F,S, \mathbf{p}, \mathbf{\Xi}^*}$ and $zero$ otherwise.   
    
\end{itemize}





\subsubsection{Computational procedure} \label{sec:3.2.6} 
This section outlines the computational procedure and associated costs for implementing an $S$-variate SDD approximation using the proposed optimal knot vector method. 
\begin{figure}[ht!]
    \centering
    \includegraphics[angle=0,scale=0.5,clip]{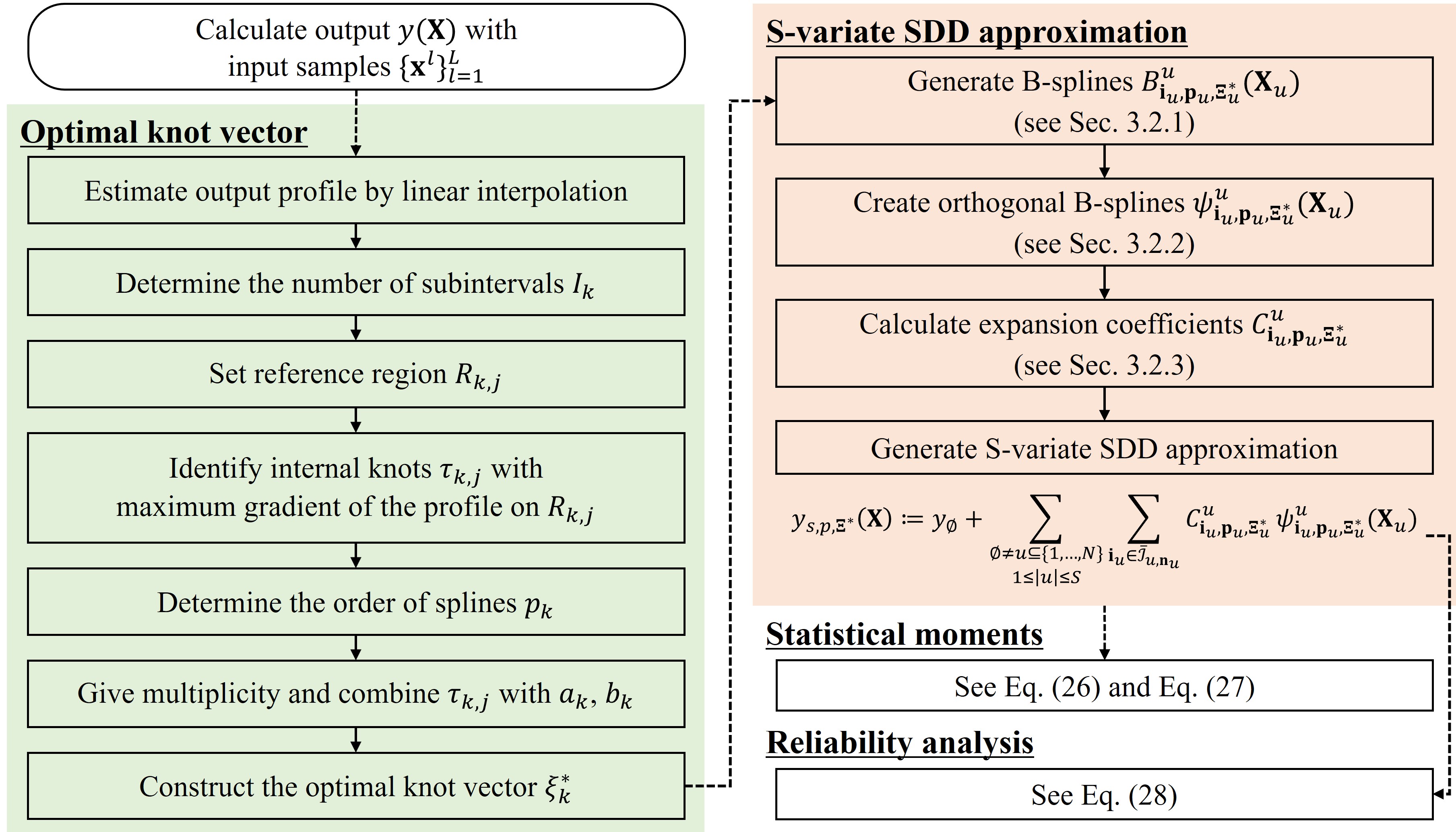}
    \caption {Flow chart of the $S$-variate SDD approximation using the proposed optimal knot vector for estimating statistical moments or failure probability of the output.}
    \label{fig15}
\end{figure}

Figure~\ref{fig15} presents a flow chart for the $S$-variate SDD approximation using the proposed optimal knot vector method. 
The procedure begins by generating output samples $y(\mathbf{X})$ from a set of input realizations $\left\{\mathbf{x}^{(l)} \right\}_{l=1}^{L}$. 
To construct an optimal knot vector, we first generate an output profile via linear interpolation to identify peak points. 
For high-dimensional problems, we define reference regions to efficiently capture multiple peaks and place internal knots at locations where the gradient of the output profile reaches its maximum within each region. 
This approach is used because the interpolated output profile is constructed from a limited number of samples, making it likely that the estimated peaks do not coincide with the true response extrema. 


Instead of relying on potentially inaccurate peak locations, the proposed method identifies points where the gradient of the interpolated profile reaches its maximum. Because these points typically correspond to regions of high nonlinearity, where the output changes rapidly with respect to the input. By placing internal knots at these locations improves the representational accuracy of the B-spline basis functions.
This gradient-based knot placement enhances the overall approximation accuracy of the SDD, particularly in capturing locally oscillating nonlinear response characteristics. 

Using the predetermined lower and upper bounds, the optimal knot vector $\mathbf{\Xi}^*$ is constructed directly, avoiding additional optimization steps and reducing computational effort. 
The right side illustrates the steps of the $S$-variate SDD approximation process, including the generation of B-splines, orthonormalization, estimation of expansion coefficients, and construction of the $S$-variate SDD. 
The hierarchical structure of SDD allows for systematic control over the approximation dimensionality by limiting the size of interacting variable subsets to at most $S$, where $1\le S \le N$. For instance, $S=1$ and $S=2$ mean univariate and bivariate approximations, respectively. 
Clearly, the required computational effort can be assessed by the number of basis functions denoted by
\begin{align}
    L_{S, \mathbf{p}, \mathbf{\Xi}^*} = 1 + \sum_{\substack{\emptyset \ne u \subseteq \{1, \ldots, N\} \\ 1 \le |u| \le S}} \prod_{k \in u} (n_k - 1) 
\le \prod_{k=1}^{N} n_k = L_{\mathbf{p}, \mathbf{\Xi}^*}
\end{align}
where $n_k$ presents the number of basis functions associated with the $k$th input dimension, $k=1,\ldots,N$. 
Compared to the conventional uniform knot placement used in standard SDD, the proposed method enables proper knot allocation by estimating high-variation regions through interpolation of the available response data. 
This leads to a better representation of localized nonlinear behavior. 
Moreover, the method does not require additional optimization techniques such as Sequential Quadratic Programming or Sequential Linear Programming, nor does it involve extra sampling beyond the existing dataset. 
Consequently, this approach maintains the computational efficiency of the SDD with a uniformly spaced knot framework while improving approximation quality.

\section{Numerical examples} \label{sec:4} 
Three numerical examples are presented to showcase the effectiveness of the proposed interpolation-based knot selection method for SDD approximation in solving UQ problems for stochastic dynamic analysis.  
The first example focuses on frequency response analysis, while the second and third address modal analysis. 
To ensure a fair comparison of accuracy across different knot vectors(uniformly spaced, randomly spaced, proposed method), the degree $p_k$, knot vector $\xi_k$, and the number of subintervals $I_k$ are kept consistent for all cases. 
Furthermore, all knot vectors are configured as $(p_k+1)-open$ knots. 
The proposed optimal knot vector method does not involve any additional optimization steps. 
As a result, the computational time required to construct the optimal knot vector is nearly identical to that of constructing uniform and random knot vectors. 
Therefore, this work does not include any time comparisons.

We additionally compare the proposed SDD with the Gaussian Process (GP) model across all examples. 
For the output $y(\mathbf{x})$, we use GP as a benchmark surrogate model, defined as
$
    y(\mathbf{x}) \sim \mathcal{GP}(m(\mathbf{x}), k(\mathbf{x}, \mathbf{x}')),
$
where $m(\mathbf{x})$ is the mean function and $k(\mathbf{x}, \mathbf{x}')$ is the covariance function between two input vectors $\mathbf{x}$ and $\mathbf{x}'$. Here, $\mathbf{x},\mathbf{x}'\in \mathbb{R}^N$ are sampled realizations of the random vector $\mathbf{X}=(X_1,\ldots,X_N)^{\intercal}$. In this study, GP is implemented using the scikit-learn library~\cite{pedregosa2011scikit}, with a zero mean function, i.e., $m(\mathbf{x}) = 0$, and a Mat\'ern kernel covariance kernel defined as 
$
k(\mathbf{x}, \mathbf{x}') = \sigma^2 \cdot \frac{2^{1-\nu}}{\Gamma(\nu)} 
\left( \frac{\sqrt{2\nu} \| \mathbf{x} - \mathbf{x}' \|}{\ell} \right)^\nu 
K_\nu\left( \frac{\sqrt{2\nu} \| \mathbf{x} - \mathbf{x}' \|}{\ell} \right),
$
where $\sigma^2 \in\mathbb{R}_0^+$ is the signal variance, $\ell\in\mathbb{R}_0^+$ is the length scale, $\nu\in\mathbb{R}_0^+$ is the smoothness parameter, $\Gamma(\cdot)$ is the gamma function, and $K_\nu(\cdot)$ is the modified Bessel function of the second kind. 
The hyperparameters ($\ell$, $\nu$) of the Mat\'ern kernel are empirically tuned to minimize the prediction error. 
The detailed values used in each case are provided in the corresponding example sections. All numerical results were obtained using Python 3.12 on an Intel i5 13400 processor with 32 GB of RAM.
%
\subsection{Example 1: frequency response analysis of a two-degree-of-freedom system} \label{sec:4.1}

\begin{figure}[ht]
    \centering
    \includegraphics[angle=0,scale=0.6,clip]{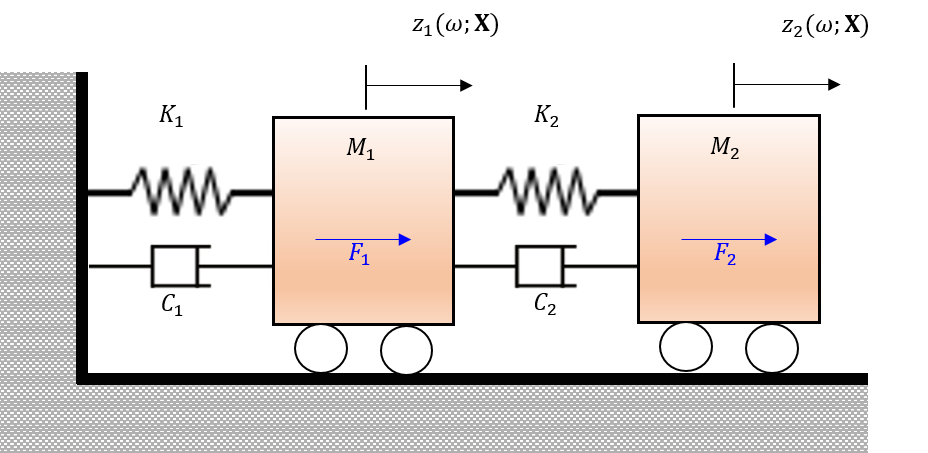}
    \caption {A two-degree-of-freedom spring--mass--damper system.~\cite{rahman2021orthogonal, jacquelin2015polynomial}}
    \label{fig3}
\end{figure}

Figure~\ref{fig3} shows a two-degree-of-freedom, proportionally damped, dynamical system with possibly random masses, damping coefficients, and spring constants $M_{1} = M_{2} = (1 + \delta_{M} X_{M})~\text{kg}, C_{1} = C_{2} = (1 + \delta_{C} X_{C})~\text{N/ms}, K_{1} = K_{2} = 15000(1 + \delta_{K} X_{K})~\text{N/m}$, respectively, where $X_M$, $X_C$, and $X_K$ are independent random variables with $\delta_M$, $\delta_C$, and $\delta_K$ representing their corresponding coefficients of variation (COVs). 
By selecting appropriate values of the COVs, various dynamical systems, whether fully deterministic (all COVs=0), fully random (all COVs$\neq$0), or a random-deterministic combination (some COVs=0), can be studied. 
Therefore, for a fully random dynamical system, the input random vector $\mathbf{X} = (X_M, X_C, X_K)^{\intercal}$.

The dynamical system is subjected to a harmonic excitation force vector $\mathbf{f}(t) = \mathbf{F}_{\omega} \exp(i\omega t)$ with the force amplitude vector $ \boldsymbol{F} = (1, 0)^{\intercal}~\text{N}
$ and angular frequency $20\pi \leq \omega \leq 70\pi ~\text{rad/s}
$. 
In terms of ordinary frequency $f = \omega /{2\pi},$, the range is $10 \leq f \leq 35~\text{Hz}$. 
From linear dynamics, the steady-state displacement response vector is $\mathbf{z}(t) = \mathbf{Z}(\omega; \mathbf{X}) \exp(i \omega t)$, where the complex-valued displacement amplitude vector $\mathbf{Z}(\omega; \mathbf{X}) = \big( Z_1(\omega; \mathbf{X}), Z_2(\omega; \mathbf{X}) \big)^\top
$ satisfies
 
\begin{align}
    \left( 
    -\omega^2 
    \begin{bmatrix}
    M_1 & 0 \\
    0 & M_2
    \end{bmatrix}
    + i \omega 
    \begin{bmatrix}
    C_1 + C_2 & -C_2 \\
    -C_2 & C_2
    \end{bmatrix}
    + 
    \begin{bmatrix}
    K_1 + K_2 & -K_2 \\
    -K_2 & K_2
    \end{bmatrix}
    \right)
    \begin{pmatrix}
    Z_1(\omega; \mathbf{X}) \\
    Z_2(\omega; \mathbf{X})
    \end{pmatrix}
    =
    \begin{pmatrix}
    1 \\
    0
    \end{pmatrix}.
    \label{eq:52}
\end{align}

Two distinct cases, one considering the randomness of the spring stiffness only $(N =1)$ and other allowing the randomness of mass, damping, and stiffness properties $(N=3)$, are studied. This example aims to perform uncertainty quantification of the displacement amplitude \( |Z_1(\omega; \mathbf{X})| \) in the frequency domain using both the SDD and GP model with a Matern kernel ($l=1.0$, $\nu=0.5$). 
First, standard deviation is estimated over a range of excitation frequencies to capture the effect of input uncertainty. Then, the failure probability is calculated using \(10^6\) Monte Carlo resamples applied to the constructed SDD and GP surrogate models.

We calculate all expansion coefficients of the SDD approximation using 10-fold cross-validated LASSO. The training set consists of five times the number of basis terms, i.e., $5 \times L_{S, \mathbf{p}, \mathbf{\Xi}}$ or $5 \times L_{S, \mathbf{p}, \mathbf{\Xi}^*}$, as a rule of thumb~\cite{lee2020practical}. 

\newcommand{\knotvectorbreak}[1]{\parbox[t]{0.8\textwidth}{#1}} 

\begin{table}[!ht] 
    \caption{Configurations of polynomial order ($p$), number of internal knots ($I$), and corresponding internal knot vectors used for SDD construction. Two types of knot spacing vectors are employed: uniformly spaced and randomly spaced, with detailed knot vectors provided for each case. These configurations are used in Example 1 to assess the impact of knot placement on approximation accuracy.}
    \centering
    \renewcommand{\arraystretch}{1.5} 
    \setlength{\tabcolsep}{3pt}       

    {\small 
    \begin{tabular}{p{1.5cm} p{1.9cm} p{5.6cm} p{5.8cm}}
    \hline 
    \multirow{2}{*}{\makecell[l]{Order of \\ polynomial}} & 
    \multirow{2}{*}{\makecell[l]{Number of \\ internal knots}} & 
    \multicolumn{2}{l}{Type of internal knots} \tabularnewline
    \cline{3-4}
     &  & Uniformly spaced knot vector & Randomly spaced knot vector \tabularnewline
    \hline 
    $p=1$ & $I=8$ & 
    $\begin{array}{l}
    \xi = \{-3, -3, -2.25, -1.5, -0.75, \\ 0, 0.75, 1.5, 2.25, 3, 3\}
    \end{array}$ & 
    $\begin{array}{l}
    \xi = \{-3, -3, -2.25, -1.25, 0.06, \\ 0.3, 1.25, 2.36, 2.28, 3, 3\}
    \end{array}$ \tabularnewline
    
     & $I=16$ & 
    $\begin{array}{l}
    \xi = \{-3, -3, -2.625, -2.25, -1.875, \\ -1.5, -1.125, -0.75, -0.375, 0, \\ 0.375, 0.75, 1.125, 1.5, 1.875, \\ 2.25, 2.625, 3, 3\}
    \end{array}$ & 
    $\begin{array}{l}
    \xi = \{-3, -3, -2.94, -2.77, -2.5, \\ -2.2, -2.15, -2.0, -1.69, -1.51, \\ -0.93, -0.49, -0.03, 0.01, 1.19, \\ 2.27, 2.71, 3, 3\}
    \end{array}$ \tabularnewline
    
    $p=2$ & $I=8$ & 
    $\begin{array}{l}
    \xi = \{-3, -3, -3, -2.25, -1.5, \\ -0.75, 0, 0.75, 1.5, 2.25, 3, 3, 3\}
    \end{array}$ & 
    $\begin{array}{l}
    \xi = \{-3, -3, -3, -2.25, -1.25, \\ 0.06, 0.3, 1.25, 2.36, 2.28, 3, 3, 3\}
    \end{array}$ \tabularnewline
    
     & $I=16$ & 
    $\begin{array}{l}
    \xi = \{-3, -3, -3, -2.625, -2.25, \\ -1.875, -1.5, -1.125, -0.75, \\ -0.375, 0, 0.375, 0.75, 1.125, \\ 1.5, 1.875, 2.25, 2.625, 3, 3, 3\}
    \end{array}$ & 
    $\begin{array}{l}
    \xi = \{-3, -3, -3, -2.94, -2.77, \\ -2.5, -2.2, -2.15, -2.0, -1.69, \\ -1.51, -0.93, -0.49, -0.03, 0.01, \\ 1.19, 2.27, 2.71, 3, 3, 3\}
    \end{array}$ \tabularnewline
    \hline 
    \end{tabular}}
    \label{tab:mat1}
\end{table}

\subsubsection{Case 1: Randomness in stiffness ($N$=1)} \label{sec:4.1.1}

The system parameters are defined as $M_1=M_2=1 \ \mathrm{kg}$, $C_1=C_2=1~\mathrm{N}/\mathrm{ms}$, $K_1=K_2=15,000(1+X) \ \mathrm{N}/\mathrm{m}$, where the random variable $X$ follows a uniform distribution over $[-3, 3]$. To assess the accuracy of the SDD approximation, univariate models with $p=1,2$ and $I=8,16$ are constructed using three types of knot vector: uniformly spaced, randomly spaced, and the proposed interpolation-based method. 
Table~\ref{tab:mat1} summarizes the corresponding knot configurations. 
Once the models are built, the standard deviation of displacement $z_1(\omega; X)$ is computed and compared against the reference solution obtained from $10^5$ crude MCS simulations. Table~\ref{tab:mat2} shows the mean errors of the standard deviation over all frequencies.

\begin{figure}[!ht]
    \centering
    \includegraphics[angle=0,scale=0.45,clip]{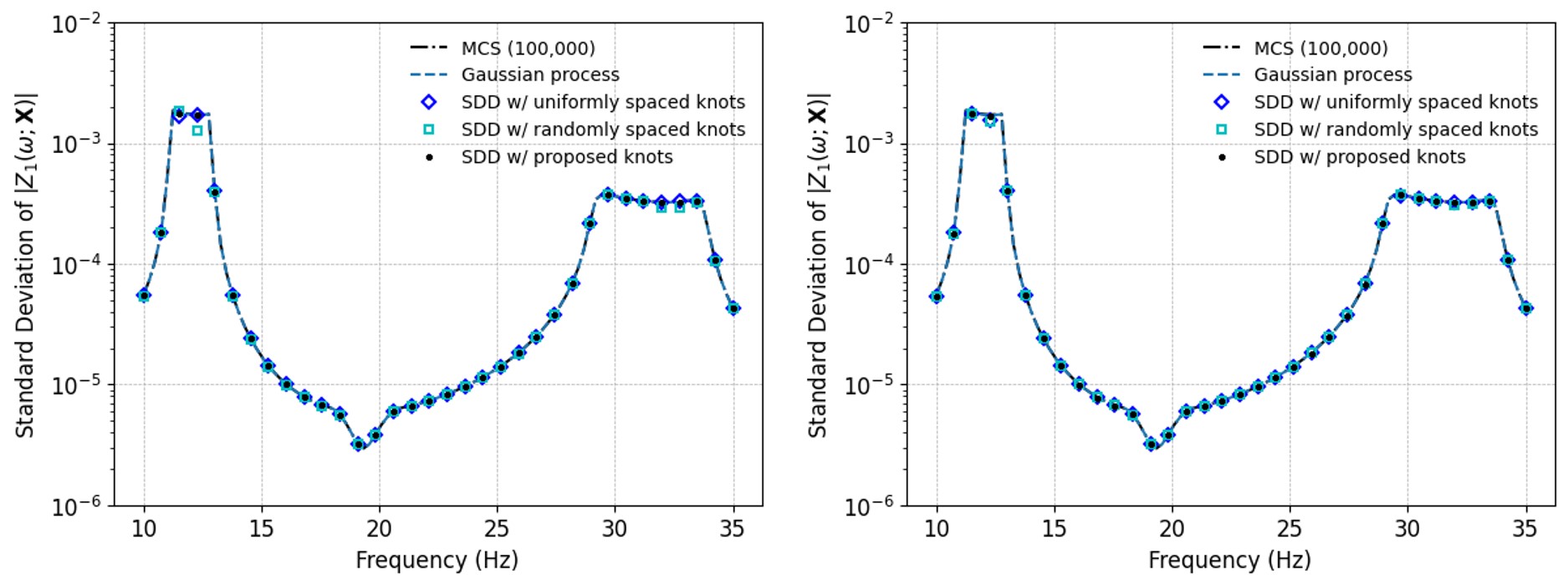}
    \caption {Standard deviation of $|Z_1(\omega;X)|$ over the frequency range from 10~Hz to 35~Hz for uniformly spaced, randomly spaced, and proposed SDD configurations compared against 100,000 Monte Carlo simulations. The results are presented for (left) $p = 1$ and $I = 16$, and (right) $p = 2$ and $I = 16$. (Case 1)}
    \label{fig4}
\end{figure}

Figure~\ref{fig4} demonstrates that the proposed SDD method exhibits good agreement with MCS across the entire frequency domain, particularly in regions of higher nonlinearity. 
Increasing both the polynomial degree $p$ and number of knots $I$ improves accuracy for all knot strategies, with the proposed method consistently yielding the lowest mean error. 
The proposed method achieves its best accuracy with $p = 2$ and $I = 16$, yielding a mean error of $0.552 \times 10^{-5}$.

\begin{table}[!ht]
    \caption{Mean errors ($\times10^{-5}$) in estimating the standard deviation of $|Z_1(\omega; \mathbf{X})|$ using SDD with various internal knot configurations and a GP model. The reference standard deviation was computed from 100,000 MCS results. Results are reported for combinations of polynomial order ($p = 1, 2$) and number of internal knots ($I = 8, 16$). The number of training samples used in each case is also listed. (Case 1)}
    \centering
    
    \renewcommand{\arraystretch}{1.5} 
    \setlength{\tabcolsep}{4pt}       
    
    {\small 
    \begin{tabular}{p{1cm} p{1cm} p{2cm} p{3.2cm} p{3.2cm} p{3.2cm} p{1cm}}
    \hline 
    \multirow{2}{*}{\textbf{$p$}} & \multirow{2}{*}{\textbf{$I$}} & \multirow{2}{*}{\makecell[l]{\textbf{Training} \\ \textbf{dataset}}} & \multicolumn{4}{l}{\textbf{Mean error} ($\times10^{-5}$)} \\

    \cline{4-7}
    & & & \textbf{Uniformly spaced} & \textbf{Randomly spaced} & \textbf{Proposed method} & \textbf{GP}\\
    \hline 
    1 & 8 & 45 & $2.115$ & $2.983$ & $1.908$ & $0.624$\\
    1 & 16 & 85 & $0.712$ & $1.489$ & $0.530$ & $0.082$ \\
    2 & 8 & 50 & $1.837$ & $1.824$ & $1.780$ & $0.583$ \\
    2 & 16 & 90 & $0.633$ & $1.064$ & $0.552$ & $0.136$ \\
    \hline 
    \end{tabular}}
    \label{tab:mat2}
\end{table}

\begin{figure}[!ht]
    \centering
    \includegraphics[angle=0,scale=0.45,clip]{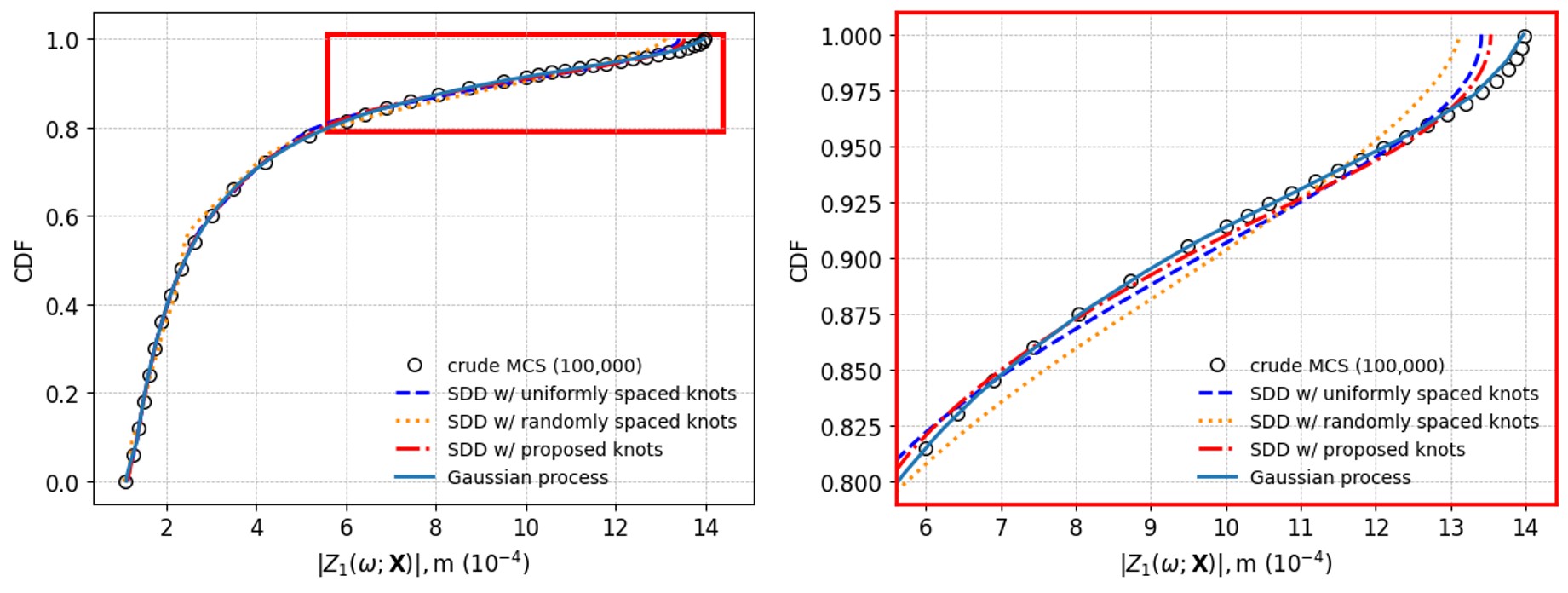}
    \caption {Cumulative probability distribution functions of $|Z_1(\omega;X)|$ at 31.465
    ~Hz, when $p=2$ and $I=16$. The results are presented for (left) displacement $|Z_1(\omega;X)|$ ranges 0 to 15, and (right) ranges 5.6 to 14.4. (Case 1)}
    \label{fig5}
\end{figure}

The second natural frequency corresponding to the mean input parameters is 31.465~Hz. To further assess dynamic behavior, cumulative distribution functions (CDFs) of \( |Z_1(\omega; X)| \) at this frequency are evaluated using $10^6$ MCS samples drawn from the constructed surrogates. 
Figure~\ref{fig5} shows that the SDD model with the proposed knot vector accurately captures the response distribution, closely matching the reference MCS results. 
However, in this low-dimensional case, the GP achieves slightly better predictive accuracy than the SDD methods. 

We assess the probability of failure at the second natural frequency (31.465~Hz) with a displacement threshold of $10^{-4}$. When any realization of \( |Z_1(\omega;X)| \) exceeds this threshold, we consider it a failure. 
Table~\ref{tab:mat3} presents the failure probabilities computed using different surrogate models, varying the polynomial order \( p \), number of internal knots \( I \), and the knot vector type (uniform, random, or the proposed optimal method). For comparison, the reference probability from crude MCS is also shown.

In all cases, the proposed SDD method demonstrates better agreement with the reference failure probability from crude MCS than both uniform and random knot vector configurations. 
In particular, the proposed method shows reduced deviation from the MCS-based benchmark, validating its robustness in failure probability estimation. 
Although GP achieves slightly higher accuracy than SDD, optimal knot selection consistently improves predictive accuracy within the SDD framework.

\begin{table}[!ht]
    \caption{Comparison of failure probabilities $\mathbb{P} \lbrack |Z_1(\omega;X)| > 10^{-4} \rbrack$ estimated using SDD with different internal knot configurations and GP model at $\omega = 31.465$\ Hz. Results are provided for various polynomial orders ($p$) and numbers of internal knots ($I$), with crude MCS results as reference (Case 1).}
    \centering
    \renewcommand{\arraystretch}{1.5} 
    \setlength{\tabcolsep}{4pt}       
    
    {\small 
    \begin{tabular}{p{1cm} p{1cm} p{3cm} p{3cm} p{3cm} p{2cm} p{2cm}}
    \hline 
    \multirow{2}{*}{\textbf{$p$}} & \multirow{2}{*}{\textbf{$I$}} &  \multicolumn{3}{l}{\textbf{Failure probability} ($\mathbb{P} \lbrack |Z_1(\omega;{X})| > 10^{-4}\rbrack $)} \\

    \cline{3-7}
    & & \textbf{Uniformly spaced} & \textbf{Randomly spaced} & \textbf{Proposed method} & \textbf{GP} & \textbf{Crude MCS} \\
    \hline 
    1 & 8 & $9.605 \times10^{-2} $ & $9.447 \times10^{-2}$ & $8.957 \times10^{-2}$ & $8.652 \times10^{-2}$ & \\
    1 & 16 & $9.411\times10^{-2}$ & $6.211\times10^{-2}$ & $9.128\times10^{-2}$ & $8.618 \times10^{-2}$ & \\
    2 & 8 & $8.083\times10^{-2}$ & $7.448\times10^{-2}$ & $8.149\times10^{-2}$ & $8.698 \times10^{-2}$ & \\
    2 & 16 & $9.344\times10^{-2}$ & $9.665\times10^{-2}$ & $9.008\times10^{-2}$ & $8.626 \times10^{-2}$ & \\
    & & & & & &  $8.586\times10^{-2}$ \\
    \hline 
    \end{tabular}}
    \label{tab:mat3}
\end{table}

\subsubsection{Case 2: Randomness in mass, damping, and stiffness (N=3)} \label{sec:4.1.2}

We define the parameters $M_1=M_2=(1+0.05X_M) \ \mathrm{kg}$, $C_1=C_2=(1+0.05X_C) \ \mathrm{N}/\mathrm{ms}=$, and $K_1=K_2=15,000(1+0.05X_K) \ \mathrm{N}/\mathrm{m}$, where random variables $\mathbf{X}=( X_M, X_C, X_K )^{\intercal}$ follow a truncated Gaussian distribution bounded in $[-3, 3]$ with standard deviations $(\delta_M=\delta_C=\delta_K=0.05)$. 
Bivariate SDD models with $p=1,2$ and $I=8$ are constructed using three different knot configurations: uniformly spaced, randomly spaced, and the proposed optimal knots. 
For both uniformly and randomly spaced knots, an identical knot vector is applied across all directions $k=1,2,3$, after which the standard deviation of $|Z_1(\omega; \mathbf{X})|$ is evaluated.

\begin{figure}[ht]
    \centering
    \includegraphics[angle=0,scale=0.48,clip]{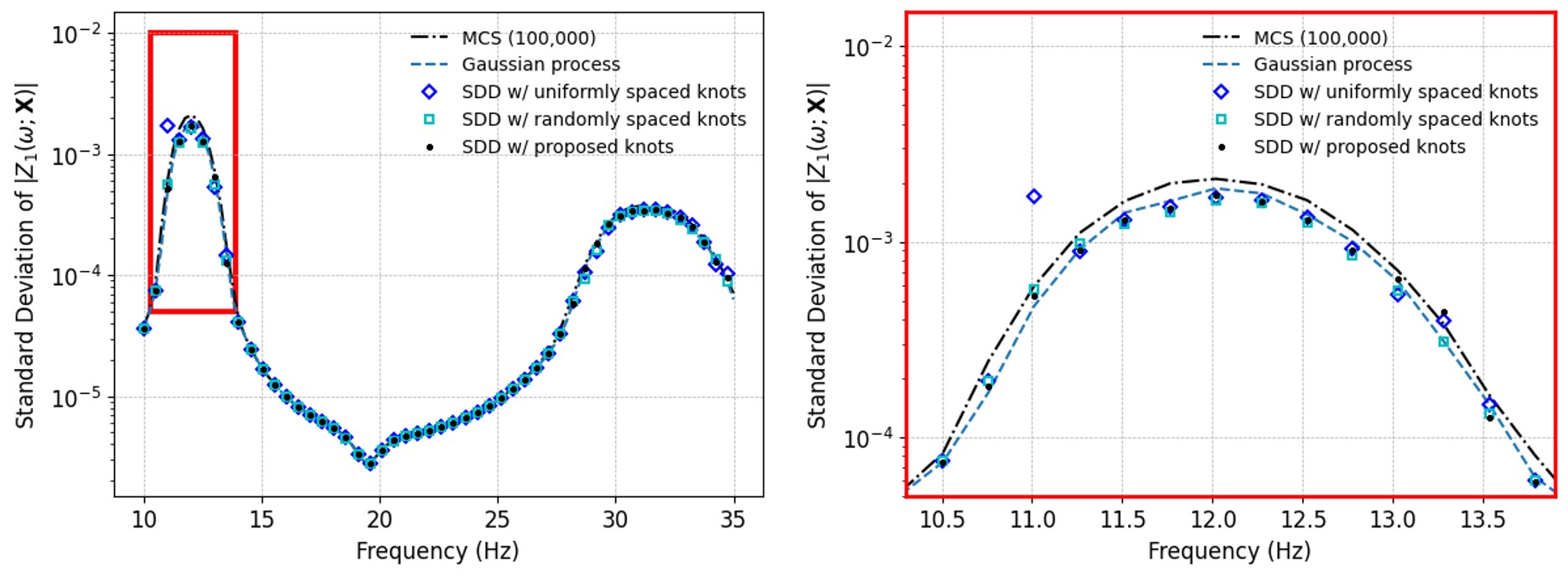}
    \caption {Standard deviation of $|Z_1(\omega; \mathbf{X})|$ over the frequency range from 10~Hz to 35~Hz for uniformly spaced, randomly spaced, and proposed SDD configurations ($p=2$, $I=8$) compared against 100,000 Monte Carlo simulations. The results are presented for frequencies (left) 10 to 35~Hz, and (right) 10.9 to 13.9~Hz. (Case 2)}
    \label{fig6}
\end{figure} 

Figure~\ref{fig6} demonstrates that the SDD model with the proposed knots, using $p=2$ and $I=8$, shows excellent agreement with the 100,000 sample MCS reference throughout the entire frequency range. 
In the 10.9--13.9~Hz region, where rapid variations occur, the proposed method accurately captures the nonlinear trend, while the uniformly spaced SDD tends to overestimate the standard deviation.

Quantitative comparison in Table~\ref{tab:mat4} confirms that the proposed method consistently achieves lower mean errors than both baseline methods across all combinations of $p$ and $I$. 
For instance, when $p=2$ and $I=16$, the mean error is reduced to $1.372 \times 10^{-5}$, compared to $1.562 \times 10^{-5}$ for uniform and $2.553 \times 10^{-5}$ for random spacing. 
Although GP still exhibits slightly better accuracy in some settings, SDD with the proposed method maintains significant advantages in flexibility.

\begin{table}[!ht]
    \caption{Comparison of mean errors in estimating the standard deviation of $|Z_1(\omega; \mathbf{X})|$ for SDD with different internal knot configurations and GP model. Results are shown for various polynomial orders ($p$), numbers of internal knots ($I$), and corresponding training dataset sizes (Case 2).}
    \centering
    \renewcommand{\arraystretch}{1.5} 
    \setlength{\tabcolsep}{4pt}       
    
    {\small 
    \begin{tabular}{p{1cm} p{1cm} p{2cm} p{3cm} p{3cm} p{3cm} p{1cm}}
    \hline 
    \multirow{2}{*}{\textbf{$p$}} & \multirow{2}{*}{\textbf{$I$}} & \multirow{2}{*}{\makecell[l]{\textbf{Training} \\ \textbf{dataset}}} & 
    \multicolumn{4}{l}{\textbf{Mean error} ($\times10^{-5}$)} \\

    \cline{4-7}
    & & & \textbf{Uniformly spaced} & \textbf{Randomly spaced} & \textbf{Proposed method} & \textbf{GP}\\
    \hline 
    1 & 8 & 1,085 & 3.917 & 4.771 & 3.852 & 2.487 \\
    2 & 8 & 1,335 & 4.141 & 3.538 & 3.111 & 2.374 \\
    1 & 16 & 4,085 & 1.319 & 2.601 & 1.311 & 1.302 \\
    2 & 16 & 4,595 & 1.562 & 2.553 & 1.372 & 1.491 \\
    \hline 
    \end{tabular}}
    \label{tab:mat4}
\end{table}

\begin{figure}[!ht]
    \centering
    \includegraphics[angle=0,scale=0.45,clip]{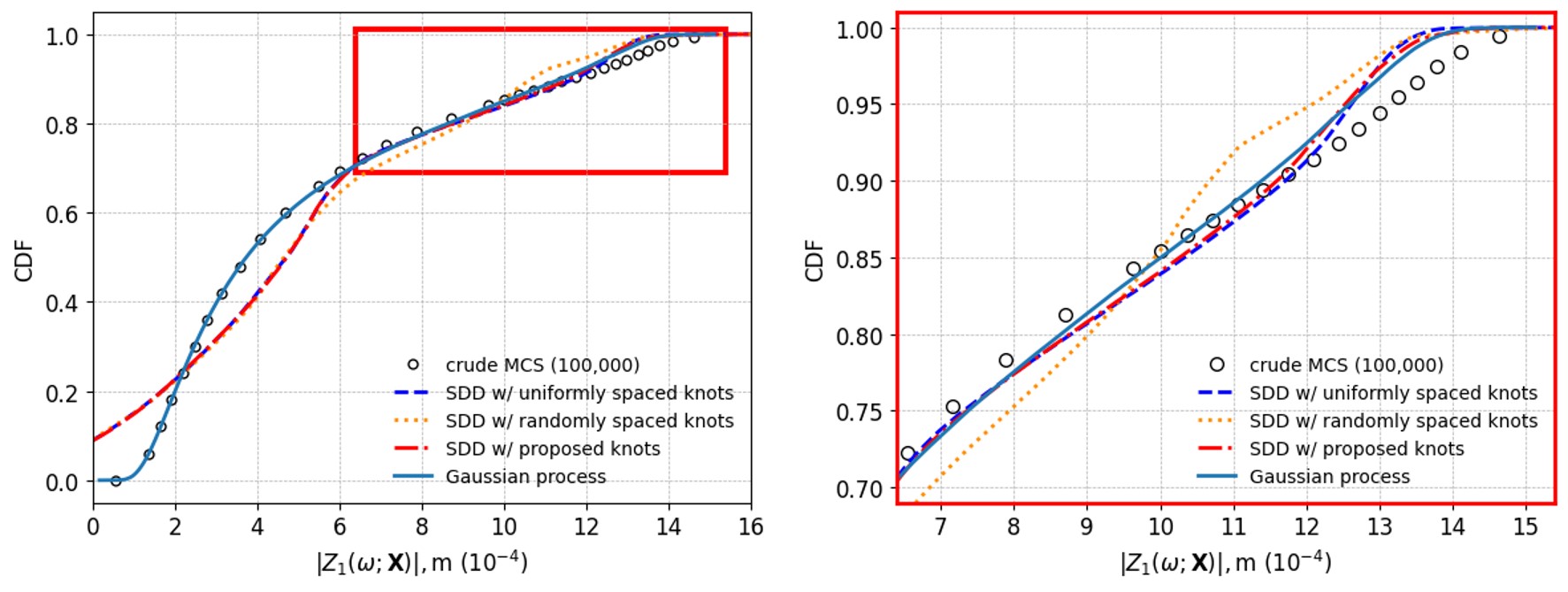}
    \caption {Cumulative probability distribution functions of $|Z_1(\omega;\mathbf{X})|$ at 31.465~Hz, when $p=2$ and $I=16$. The results are presented for (left) displacement $|Z_1(\omega;\mathbf{X})|$ ranges 0 to 16, and (right) ranges 6.4 to 15.4 (Case 2)}
    \label{fig7}
\end{figure}

Figure~\ref{fig7} presents the cumulative distribution functions (CDF) of $|Z_1(\omega; \mathbf{X})|$ at the second natural frequency (31.465~Hz), computed using $10^6$ resamples from each surrogate. 
The failure probability is defined as the probability of the response that exceeds a threshold value of $10^{-4}$. 
Table~\ref{tab:mat5} compares the failure probabilities computed by each method against the reference values from crude MCS. 
The proposed SDD model demonstrates excellent agreement with the MCS benchmark, especially when compared to randomly spaced SDD. This confirms its effectiveness in capturing rare-event behavior in complex uncertainty conditions.

\begin{table}[!ht]
    \caption{Comparison of failure probability estimates for $|Z_1(\omega; \mathbf{X})|$ at 31.465\ Hz across different knot configurations and GP model, benchmarked against crude MCS results (Case 2). Failure occurs when displacement exceeds $10^{-4}$. Results are reported for polynomial orders $p=1, 2$ and internal knots $I=8, 16$.}
    \centering
    \renewcommand{\arraystretch}{1.5} 
    \setlength{\tabcolsep}{4pt}       
    
    {\small 
    \begin{tabular}{p{1cm} p{1cm} p{3cm} p{3cm} p{3cm}  p{2cm} p{2cm}}
    \hline 
    \multirow{2}{*}{\textbf{$p$}} & \multirow{2}{*}{\textbf{$I$}} &  \multicolumn{3}{l}{\textbf{Failure probability} ($\mathbb{P} \lbrack |Z_1(\omega;\mathbf{X})| > 10^{-4}\rbrack $)} \\

    \cline{3-7}
    & & \textbf{Uniformly spaced} & \textbf{Randomly spaced} & \textbf{Proposed method} & \textbf{GP} & \textbf{crude MCS} \\
    \hline 
    1 & 8 & $1.130 \times10^{-1} $ & $0.982 \times10^{-1}$ & $1.139 \times10^{-1}$ & $1.381 \times10^{-1}$ & -\\
    1 & 16 & $1.604\times10^{-1}$ &  $1.211\times10^{-1}$ & $1.562\times10^{-1}$ & $1.482 \times10^{-1}$ & - \\
    2 & 8 & $1.534\times10^{-1}$ & $1.146\times10^{-1}$ & $1.404\times10^{-1}$ & $1.467 \times10^{-1}$ & -\\
    2 & 16 & $1.611\times10^{-1}$ & $1.457\times10^{-1}$ & $1.594\times10^{-1}$ & $1.505 \times10^{-1}$ & - \\
    -& -& -& -& -& -&  $1.458\times10^{-1}$ \\
    \hline 
    \end{tabular}}
    \label{tab:mat5}
\end{table}

\subsection{Example 2: Modal analysis of a turbo fan jet engine (N=5)} \label{sec:4.2}

\begin{figure}[!ht]
    \centering
    \includegraphics[angle=0,scale=0.35,clip]{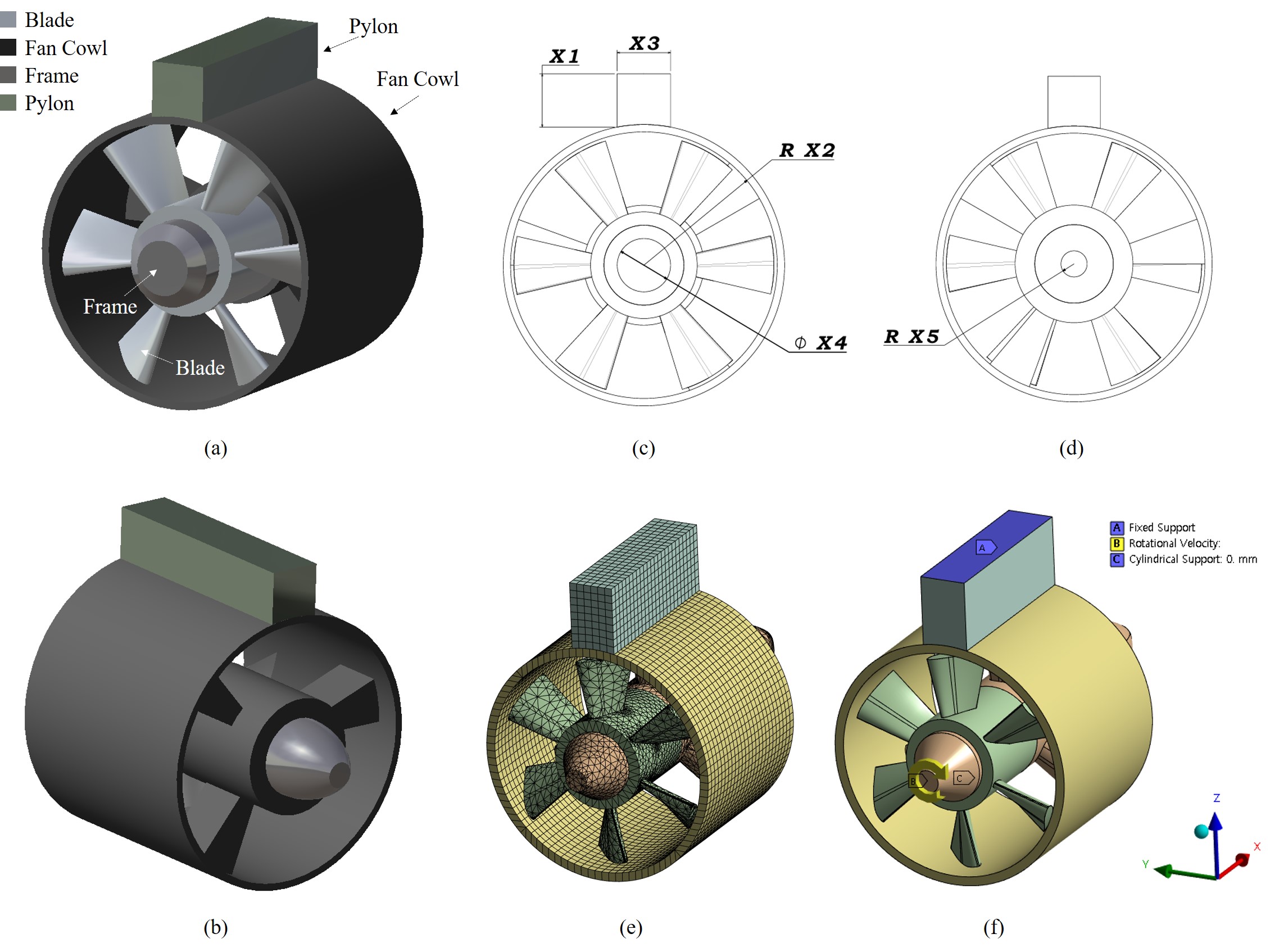}
    \caption {The 3D model of turbofan jet engine with detailed views highlighting the components and material composition (a) front view and (b) back view (Example 2); A CAD model of the turbofan jet engine; (c) front view of the engine with four input random variables and (d) back view of the engine with single input random variables; An finite element analysis of the engine; (e) a mesh consisting of 71,406 nodes and 29,995 elements in the nominal model (f) 5000 rpm rotational velocity and boundary conditions, including fixed support and cylindrical support.}
    \label{fig8}
\end{figure}

\hspace{1em} The second example demonstrates the application of the proposed SDD method to quantify the uncertainty in the first natural frequency of a turbofan jet engine. Figure~\ref{fig8} illustrates that the engine is mounted on the wing through a pylon, and the rotating internal blades generate thrust. As variations in the structural design variables of the engine induce variability in its dynamic response, accurate natural frequency prediction is  essential to prevent resonance and ensuring reliability.

Table~\ref{tab:mat6} summarizes that five independent design parameters follow uniform distribution with a 5\% coefficient of variance (COV). 
These parameters are used to predict the first natural frequency of the turbofan jet engine. 
The vibration analysis is conducted using ANSYS (Version 2024 R1\cite{ansys2024r1}) with the boundary conditions illustrated in Figure~\ref{fig8}. 
The pylon's upper surface is attached to the aircraft wing using a fixed support. 
To account for blade rotation, a cylindrical support and rotational velocity of 5000 [rpm] are applied for vibration analysis. The nominal model's mesh consists of 71,406 nodes and 29,995 elements. 
All components are modeled using structural steel, with material properties detailed in Appendix~\ref{sec:appx2}.

\begin{table}[!ht]
    \caption{Definition of five independent input random variables used for uncertainty quantification (UQ) of the turbofan jet engine's natural frequencies (Example 2). Each variable follows a uniform distribution with a coefficient of variation (COV) of 5\%. The table provides the mean, lower, and upper bounds for each input.}
    \centering
    \renewcommand{\arraystretch}{1.5} 
    \setlength{\tabcolsep}{5pt}       
    
    {\small 
    \begin{tabular}{p{2.5cm} p{2cm} p{3.5cm} p{2.5cm} p{2.5cm}}
    \hline 
    \multirow{2}{*}{\makecell[l]{\textbf{Random} \\ \textbf{variable}}}& 
    \multirow{2}{*}{\makecell[l]{\textbf{Mean} \\ \textbf{(mm)}}} & 
    \multirow{2}{*}{\makecell[l]{\textbf{Coefficient of} \\ \textbf{variation (COV$_{i}$$^{a}$, \%)}}} & 
    \multirow{2}{*}{\makecell[l]{\textbf{Lower} \\ \textbf{boundary(mm)}}} & 
    \multirow{2}{*}{\makecell[l]{\textbf{Upper} \\ \textbf{boundary(mm)}}} \\
    \\[-1.5ex] 
    \hline 
    ${X}_{1}$ & 40.0 & 5 & 38.000 & 42.000 \\
    ${X}_{2}$ & 105.5 & 5 & 102.335 & 108.665 \\
    ${X}_{3}$ & 20.0 & 5 & 19.400 & 20.600 \\
    ${X}_{4}$ & 20.0 & 5 & 19.400 & 20.600 \\
    ${X}_{5}$ & 10.0 & 5 & 9.500 & 10.500 \\
    \hline 
    \end{tabular}}
    \label{tab:mat6}
\end{table}

To construct surrogate models, we generate 400 simulation data points using the Latin hypercube sampling method. 
Univariate SDD models are trained using three different knot configurations (uniformly spaced, randomly spaced, and the proposed method) with $p=1$ and $I=42$. 
For comparison, a GP model with a Matern kernel ($l=1.0$, $\nu=1.5$) is also trained on the same dataset. 
The prediction accuracy of each surrogate model is evaluated by comparing its output with 2,000 MCS results.

\begin{figure}[!ht]
    \centering
    \includegraphics[angle=0,scale=0.65,clip]{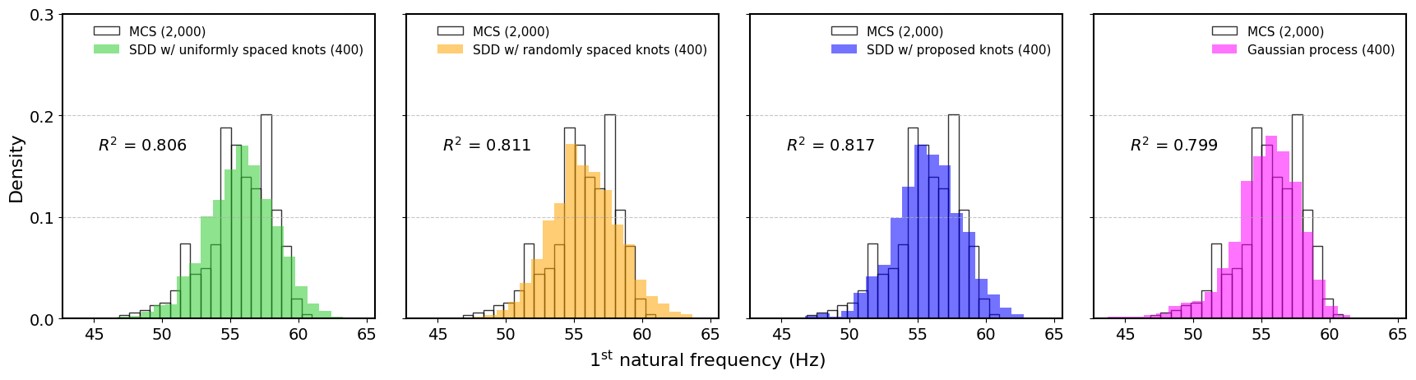}
    \caption {Compare the distribution of natural frequency at $1^{\mathrm{st}}$ SDD with uniformly spaced knots, randomly spaced knots, proposed knots and Gaussian process model trained by 400 simulations with 2,000 MCS results}
    \label{fig9}
\end{figure}

The predictive performance is evaluated using the coefficient of determination ($R^2\in [0,1]$), which quantifies the agreement between predicted and observed values, and the variance of the natural frequency. 
Figure~\ref{fig9} shows the predicted distributions compared with MCS results. The proposed SDD model achieves the highest $R^2$ value of 0.817, compared to 0.811 and 0.806 for randomly and uniformly spaced ones, respectively. The GP model yields an $R^2$ of 0.799, indicating a 2.25\% improvement by the proposed method over GP. 
While the GP model exhibited more accuracy in Example~1, the SDD model demonstrates greater predictive accuracy for computational simulation-based uncertainty quantification.

Table~\ref{tab:mat7} presents the stochastic properties of each model with crude MCS results. The proposed SDD model provides the closest estimate to MCS in variance, with the lowest relative standard deviation error (0.699\%) among all models. 
This confirms the robustness and practicality of the proposed knot selection method for modeling structural dynamics with random variables.

\begin{table}[!ht]
    \caption{Comparison of mean, variance, and relative standard deviation error ($\sigma_{\text{rel}}$) of the first natural frequency predicted by SDD models using uniformly spaced, randomly spaced, and proposed optimal knots, as well as a Gaussian Process (GP) model. All surrogate predictions are benchmarked against results from 2,000 crude MCS results. (Example 2)}
    \centering
    \renewcommand{\arraystretch}{1.5} 
    \setlength{\tabcolsep}{6pt}       

    {\small
    \begin{tabular}{l c c c c c}
    \hline
    \textbf{Properties} & 
    \textbf{Uniformly spaced} & 
    \textbf{Randomly spaced} & 
    \textbf{Proposed method} & 
    \textbf{GP} & 
    \textbf{Crude MCS} \\
    \hline
    Mean & 55.757 & 55.807 & 55.752 & 55.295 & 55.621 \\
    Variance & 6.123 & 6.281 & 5.965 & 6.144 & 6.007 \\
    \makecell[l]{Error of std \\ ($\sigma_{\text{rel}}$, \%)} & 1.931 & 4.561 & 0.699 & 2.281 & -- \\
    

    \hline
    \end{tabular}}
    \label{tab:mat7}
\end{table}

\subsection{Example 3: Modal analysis of lower control arm (N=10)} \label{sec:4.3}

\begin{figure}[!ht]
    \centering
    \includegraphics[angle=0,scale=0.4,clip]{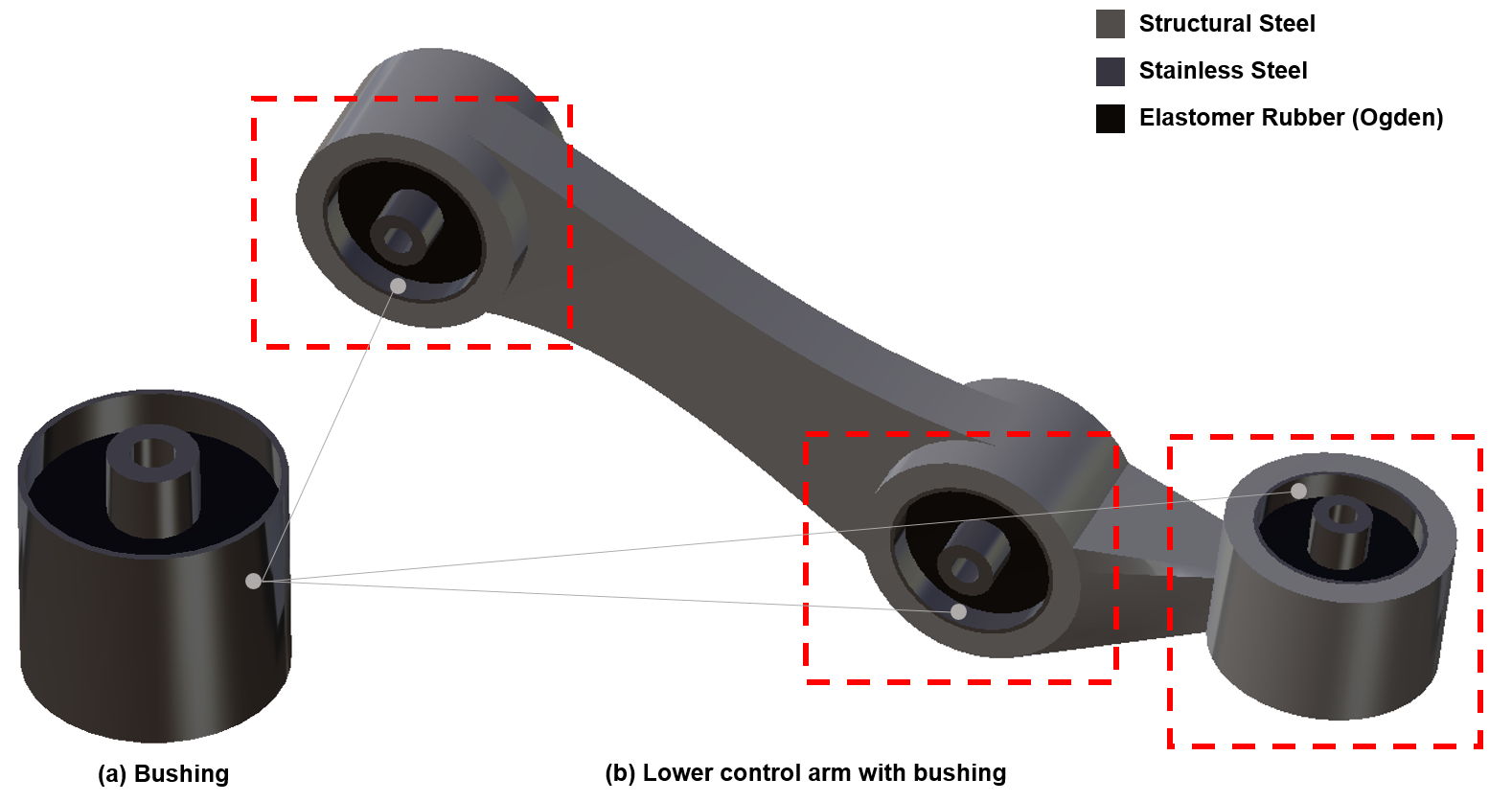}
    \caption {The 3D model of (a) a bushing and (b) a lower control arm incorporating three bushings, with detailed views highlighting the bushing components and material composition (Example 3).}
    \label{fig10}
\end{figure}

This example addresses a high-dimensional UQ problem involving the modal analysis of a lower control arm with three rubber bushings. 
Figure~\ref{fig10} illustrates the lower control arm, a critical component of the vehicle's suspension system that links the vehicle body to the wheel. Its primary function is to absorb shocks and facilitate controlled motion of the suspension assembly.
Due to its complex geometry and multiple interfaces, manufacturing variations or tolerances significantly affects the dynamic characteristics.

\begin{figure}[!ht]
    \centering
    \includegraphics[angle=0,scale=0.55,clip]{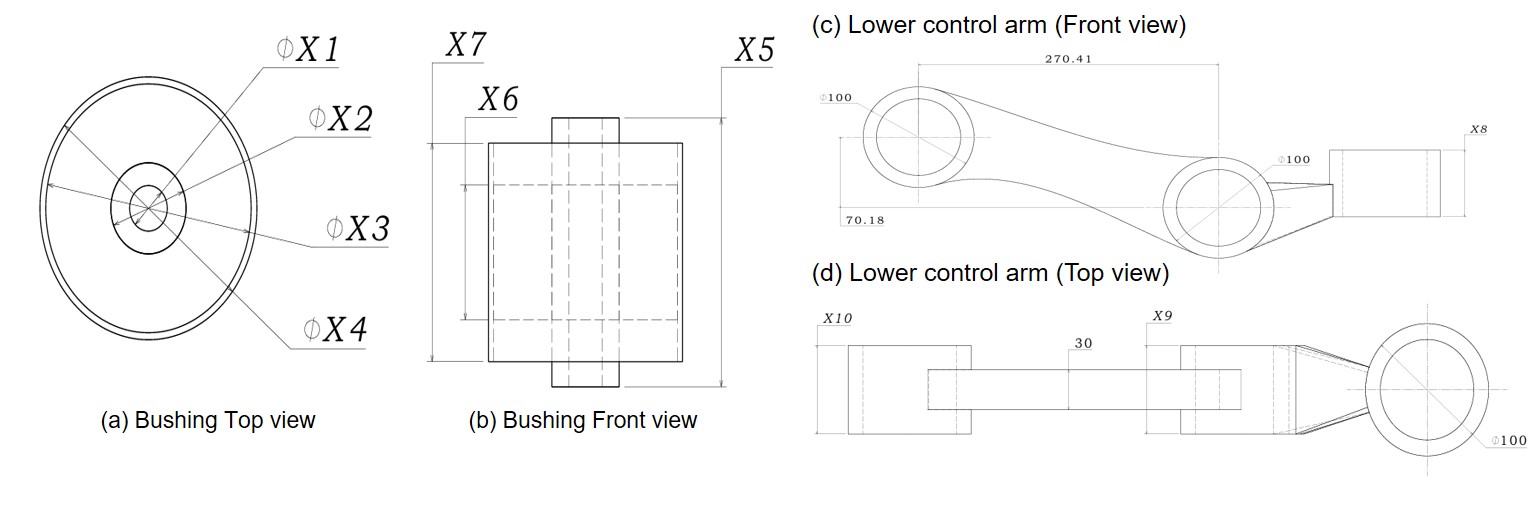}
    \caption {A CAD model of the lower control arm and bushing (Example 3); (a) top view of bushing with four input random variables (${X}_{1}$, ${X}_{2}$, ${X}_{3}$, ${X}_{4}$); (b) front view of bushing with three input random variables (${X}_{5}$, ${X}_{6}$, ${X}_{7}$); (c) front view of lower control arm with single input random variable (${X}_{8}$); (d) top view of lower control arm with two input random variables (${X}_{9}$, ${X}_{10}$)}
    \label{fig11}
\end{figure}

The objective is to evaluate the statistical moments and distribution of the natural frequencies using the proposed SDD model under high-dimensional uncertainty. The system contains 10 independent random variables, each representing geometric tolerances of the bushing and control arm. 
Table~\ref{tab:mat10} summarizes and Figure~\ref{fig11} illustrates the input parameters, which follow uniform distributions with a coefficient of variation of 5\%. 

\begin{table}[!ht]
    \caption{Definition of the 10 input random variables for uncertainty quantification of the natural frequencies in the lower control arm system. Each variable represents a geometric tolerance in the control arm or bushing, modeled as an independent uniform distribution with 5\% coefficient of variation (COV). Boundaries are calculated based on the specified mean and COV. (Example 3)}
    \centering
    \renewcommand{\arraystretch}{1.5} 
    \setlength{\tabcolsep}{5pt}       
    
    {\small 
    \begin{tabular}{p{2.5cm} p{2cm} p{3.5cm} p{2.5cm} p{2.5cm}}
    \hline 
    \multirow{2}{*}{\makecell[l]{\textbf{Random} \\ \textbf{variable}}}& 
    \multirow{2}{*}{\makecell[l]{\textbf{Mean} \\ \textbf{(mm)}}} & 
    \multirow{2}{*}{\makecell[l]{\textbf{Coefficient of} \\ \textbf{variation (\%)}}} & 
    \multirow{2}{*}{\makecell[l]{\textbf{Lower} \\ \textbf{boundary(mm)}}} & 
    \multirow{2}{*}{\makecell[l]{\textbf{Upper} \\ \textbf{boundary(mm)}}} \\
    \\[-1.5ex] 
    \hline 
    ${X}_{1}$ & 13.0 & 5 & 12.350 & 13.650 \\
    ${X}_{2}$ & 26.0 & 5 & 24.700 & 27.300 \\
    ${X}_{3}$ & 40.0 & 5 & 38.000 & 42.000 \\
    ${X}_{4}$ & 63.0 & 5 & 59.850 & 66.150 \\
    ${X}_{5}$ & 20.0 & 5 & 19.000 & 21.000 \\
    ${X}_{6}$ & 75.0 & 5 & 71.250 & 78.750 \\
    ${X}_{7}$ & 32.5 & 5 & 30.875 & 34.125 \\
    ${X}_{8}$ & 37.5 & 5 & 35.625 & 39.375 \\
    ${X}_{9}$ & 37.5 & 5 & 35.625 & 39.375 \\
    ${X}_{10}$ & 75.0 & 5 & 71.250 & 78.750 \\
    \hline 
    \end{tabular}}
    \label{tab:mat10}
\end{table}

\begin{figure}[!ht]
    \centering
    \includegraphics[angle=0,scale=0.5,clip]{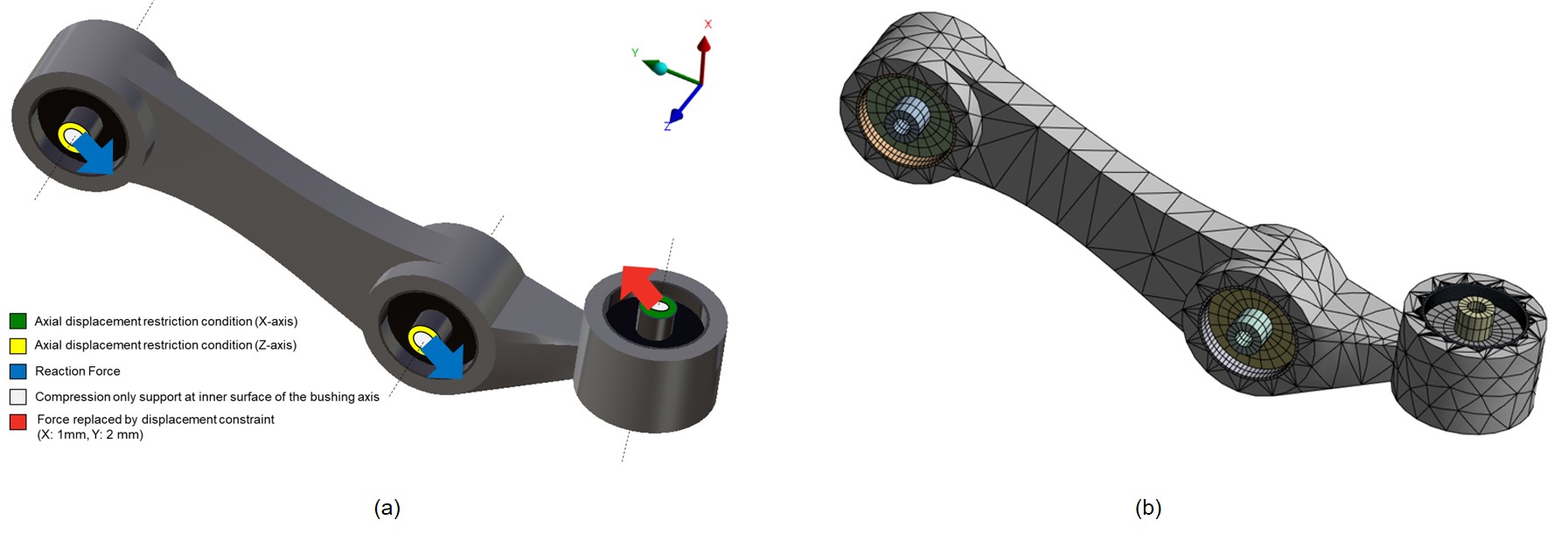}
    \caption {An finite element model of the lower control arm with bushings (Example 3); (a) loading and boundary conditions, including axial displacement conditions along the x- and z-axes, compression-only support, and force replaced by displacement constraints (x-axis: 1 [mm], y-axis: 2 [mm]); (b) a tetrahedral mesh (the lower control arm) and 
 a hexahedral mesh (the bushing) consisting of 154,855 nodes and 41,286 elements in the nominal model.}
    \label{fig12}
\end{figure}

A finite element analysis was conducted using ANSYS 2024 R1. Material properties used in the analysis are summarized in Appendix~\ref{sec:appx2}. The lower control arm is modeled using structural steel, while the outer casing of the bushing is made of stainless steel. To accurately capture the hyperelastic behavior of the rubber inside the bushings, both the Ogden model and the Prony series are employed. The Ogden model accounts for the nonlinear elastic response of rubber materials, whereas the Prony series describes viscoelastic effects under dynamic loading. These models were previously validated and are applied in this study to ensure realistic representation of material responses under vibration~\cite{lee2019prediction}. The nominal model consists of 154,855 nodes and 41,286 elements, utilizing hexahedral and tetrahedral meshes for the bushings and control arm, respectively. Appropriate boundary and loading conditions were applied, as shown in Figure~\ref{fig12}.

Univariate SDD models with $p=1$ and $I=20$ across all 10 coordinate directions $(k=1, \dots,10)$ were constructed using uniformly spaced, randomly spaced, and proposed optimal knot vectors. Each SDD model comprised 201 basis functions, and the expansion coefficients were computed using standard least squares with 401 training samples generated via Latin hypercube sampling. For comparison, a GP model was also trained using the same dataset. The GP employed a Metern kernel with a length scale of 1.0 and a smoothness parameter $\nu=2.5$.

Table~\ref{tab:mat11} presents the relative variance errors of the first and second natural frequencies estimated by SDD and GP models, benchmarked against the results from 2,000 crude MCS results. The proposed optimal knot selection method consistently achieved the lowest relative errors across all modes. Specifically, for the first mode, the proposed method reduced the variance error to 2.88\%, which corresponds to a 76.6\% reduction compared to the uniformly spaced knots (12.31\%) and an 81.1\% reduction compared to the randomly spaced knots (15.27\%). For the second mode, a similar trend was observed, with the proposed method yielding an exceptionally low error of 0.10\%, outperforming all other configurations.

\begin{table}[!ht]
    \caption{Comparison of mean, variance, and relative variance error ($\sigma_{\text{rel}}$) of the first and second natural frequencies estimated by various SDD models and GP, benchmarked against crude MCS results with 2,000 samples ($p=1$, $I=20$, Example 3).}
    \centering
    \renewcommand{\arraystretch}{1.5} 
    \setlength{\tabcolsep}{8pt}       
    
    {\small 
    \begin{tabular}{p{4.5cm} p{1cm} p{1.5cm} p{1.8cm} p{1cm} p{1.5cm} p{1.8cm}}
    \hline 
    \multirow{2}{*}{\textbf{\makecell[t]{Model}}} & 
    \multicolumn{3}{l}{\textbf{$1^{\mathrm{st}}$ mode}} &
    \multicolumn{3}{l}{\textbf{$2^{\mathrm{nd}}$ mode}} \\
    \cline{2-4} \cline{5-7} 
     & mean & variance & Error of std ($\sigma_{\text{rel}}$, \%) & mean & variance & Error of std ($\sigma_{\text{rel}}$, \%) \\
    \hline 
    SDD w/ uniformly spaced knots & 30.657 & 1.154 & 12.310 & 37.053 & 0.912 & 6.653 \\
    SDD w/ randomly spaced knots& 30.661 & 1.115 & 15.274 & 37.057 & 0.856 & 12.385 \\
    SDD w/ proposed knots & 30.658 & 1.354 & 2.888 & 37.054 & 0.978 & 0.102 \\
    GP & 30.396 & 1.246 & 5.319 & 36.830 & 1.133 & 15.967 \\
    Crude MCS (2,000) & 30.656 & 1.316 & -- & 37.053 & 0.977 & -- \\
    \hline 
    \end{tabular}}
    \label{tab:mat11}
\end{table}

\begin{figure}[!ht]
    \centering
    \includegraphics[angle=0,scale=0.65,clip]{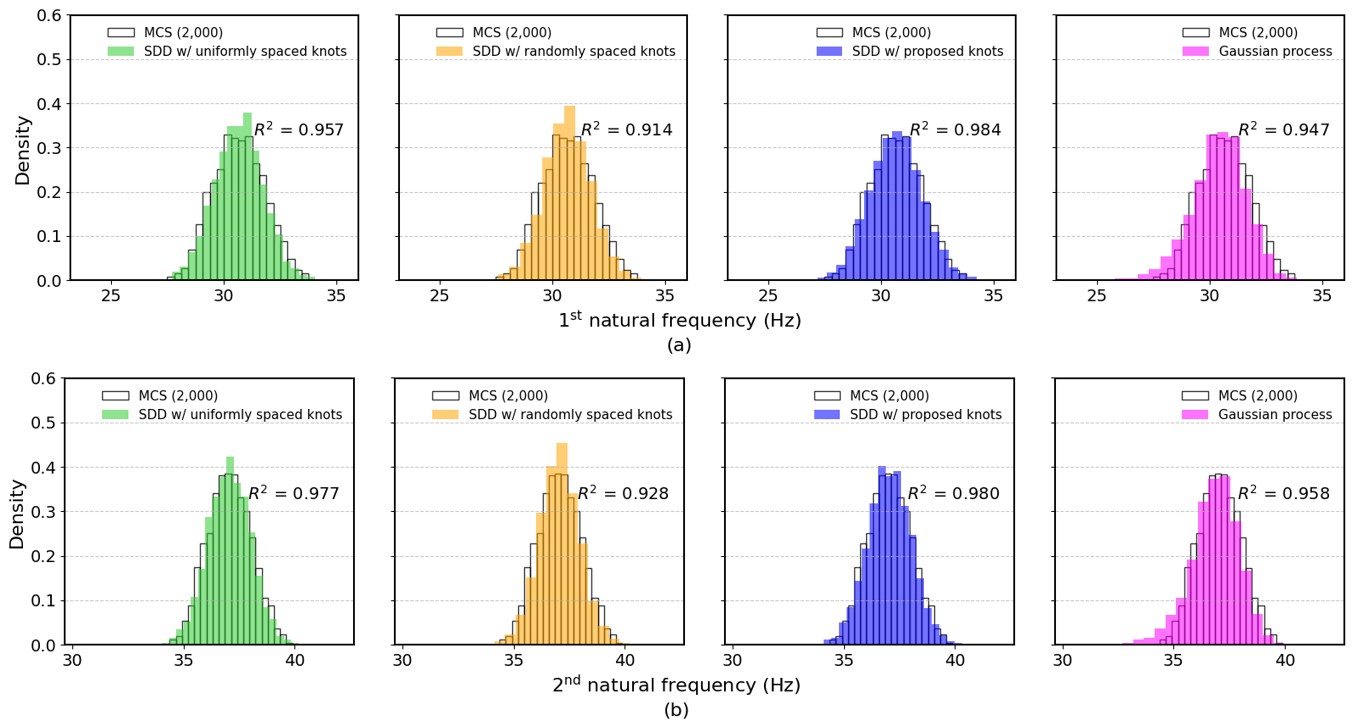}
    \caption {Compare the distribution of natural frequency at $1^{\mathrm{st}}$ and $2^{\mathrm{nd}}$ uniformly spaced, randomly spaced, and proposed SDD with 2,000 Monte Carlo Simulation results (Example 3)}
    \label{fig13}
\end{figure}

To further assess model accuracy, one million input samples were generated based on the ten random variables, and the trained surrogate models were evaluated on this larger set. As illustrated in Figure~\ref{fig13}, the SDD model with the proposed knots closely approximated the distribution of both the first and second natural frequencies about MCS. The $R^2$ value for the first mode reached 0.984, indicating a 7.7\% improvement over the randomly spaced knot model (0.914) and a 2.8\% improvement over the uniformly spaced knot model (0.957). A similar performance gain was observed for the second mode. These results confirm that the proposed knot method not only improves variance accuracy but also preserves distributional characteristics with fewer training samples (401 simulation results), compared to the 2,000 MCS results. The number of basis functions and the size of the training dataset are fixed for all knot configurations. As a result, the computational cost required to train the surrogate models remains unchanged. Therefore, the improved accuracy achieved by the proposed method stems solely from the more effective placement of internal knots, not from increased data or model complexity. This highlights the SDD with the proposed method as a statistically efficient and scalable approach for high-dimensional UQ in structural dynamics.

\section{Conclusions} \label{sec:5} 
This study proposed a novel optimal spline dimensional decomposition (SDD) method that incorporates an optimal knot vector for uncertainty quantification in dynamical systems. While B-spline basis functions are effective for modeling non-smooth responses, their performance is highly sensitive to the placement of internal knots. To address this issue, we developed an interpolation-based knot selection method that avoids additional optimization. The method first approximates the input-output data profile using linear interpolation. It then identifies reference regions that exhibit high gradient variation, which are particularly important in high-dimensional problems with oscillatory responses. Within each region, the point of maximum gradient is selected as an internal knot, thereby constructing an optimal knot vector tailored to the stochastic characteristics of the response. This optimal knot vector is then embedded into the spline basis to develop an optimal SDD model.

The effectiveness of the proposed method is validated through three numerical examples, which include frequency and modal analyses with up to ten random variables. Compared to uniformly and randomly spaced knots, the proposed approach consistently yielded lower variance errors, and estimated cumulative distribution functions (CDFs) are in good agreement with MCS. The proposed method proved its effectiveness in solving higher-dimensional scenarios that often apply to industrial structures such as a turbofan jet engine and lower control arm. While the proposed SDD model showed slightly lower accuracy than Gaussian Process (GP) models in low-dimensional cases, it outperformed GP in the case of increased input dimension. These results demonstrate the method's capability to capture strong non-linearities using appropriately placed knots, even with limited data. Consequently, the method successfully estimated failure probability estimates that are in close agreement with MCS.

 However, the current approach may have limited accuracy with uncertain response shapes as it estimates input-output profiles using linear interpolation. In addition, the method cannot currently predict frequency-domain responses in a unified manner. Incorporating higher-order interpolation schemes (e.g., quadratic or cubic) can improve accuracy. Future work will address these limitations by exploring advanced interpolation methods and integrating intrusive dynamical system modifications to enable full-spectrum frequency predictions. Ultimately, the proposed method's potential in design optimization will be further evaluated by embedding reliability indices within RBDO frameworks.

\section*{Acknowledgment} \label{sec:6} 

\noindent\textbf{\textit{Funding}}
 Y. Kim was in part financially supported by Basic Science Research Program through the National Research Foundation of Korea (NRF) funded by the Ministry of Education (RS-2024-00464478). D. Lee was in in part financially supported by the National Research Foundation of Korea (NRF) grant funded by the Korea government (MSIT) (RS-2025-00560781).

\begin{appendices}
\section{B-splines \label{sec:appx1}}

\subsection{Knot vector \label{sec:appx1.1}}
The following introduces two characteristics of the knots referred as Cottrell et al~\cite{cottrell2009isogeometric, de1972calculating}.
    
    \begin{enumerate}
    \item A knot vector is a non-decreasing sequence of real numbers defined within the bounds $\lbrack a_k,b_k\rbrack$. An example is $\xi = \lbrace1,2,3,4,5\rbrace$.

    \item Knot multiplicity indicates that individual knots can appear not only once but multiple times. The range of multiplicity is $1\leq m_{k,j_k} \leq p_k+1$. In particular, a bounded knot with $p_k+1$ is referred to as $(p_k+1)\text{-open}$. If it also has one internal knot, it is referred to as $(p_k+1)\text{-open with simple knots}$. Only $(p_k+1)\text{-open}$ knot vector is selected in this work. When knot multiplicity is present, the knot vector is defined as
        \begin{align}      
        \xi_k = \{ a_k = \underbrace{\xi_{k,1}, \dots, \xi_{k,1}}_{m_{k,1} \text{ times}}, 
        \dots, \underbrace{\xi_{k,n_k+p_k+1}, \dots, \xi_{k,n_k+p_k+1}}_{m_{k,n_k+p_k+1} \text{ times}} = b_k \}.
        \label{eq:56}
        \end{align}
    \end{enumerate}

    \subsection{Recursive formula \label{sec:appx1.2}}
    For a given knot vector $\xi_k$ and polynomial order $p_k$, univariate B-splines $B_{i_k,p_k,\xi_k}^k (x_k)$ are defined in Eq.~\eqref{eq:57} and ~\eqref{eq:58}, based on the order and derived using the methods of Cox and de Boor~\cite{cox1972numerical, de1972calculating, piegl2012nurbs}. Those equations presented $zero$-order function($p=0$) and $higher$-order function ($p=1,...,N$) 

    \begin{align}
    B_{i_k,0,\xi_k}^k (x_k) =
    \begin{cases}
        1, & \xi_{k,i_k} \leq x_k < \xi_{k,i_k+1}, \\
        0, & \text{otherwise},
    \end{cases}
    \label{eq:57}
    \end{align}

    \begin{align}
    B_{i_k,p_k,\xi_k}^k (x_k) = 
    \frac{x_k \space-\space \xi_{k,i_k}}{\xi_{k,i_k+p_k}\space-\space \xi_{k,i_k}} \space B_{i_k,p_k-1,\xi_k}^k (x_k) \space + \frac{\xi_{k,i_k+p_k+1} \space-\space  x_k}{\xi_{k,i_k+p_k+1}\space-\space \xi_{k,i_k+1}} B_{i_k+1,p_k-1,\xi_k}^k (x_k), 
    \label{eq:58}
    \end{align}

    where $1 \le k \le N$, $1 \le i_k \le n_k$, $1 \le p_k < \infty$, $0/0$ is considered as $zero$.

    \begin{figure*}
    \begin{center}
    \includegraphics[angle=0,scale=0.6,clip]{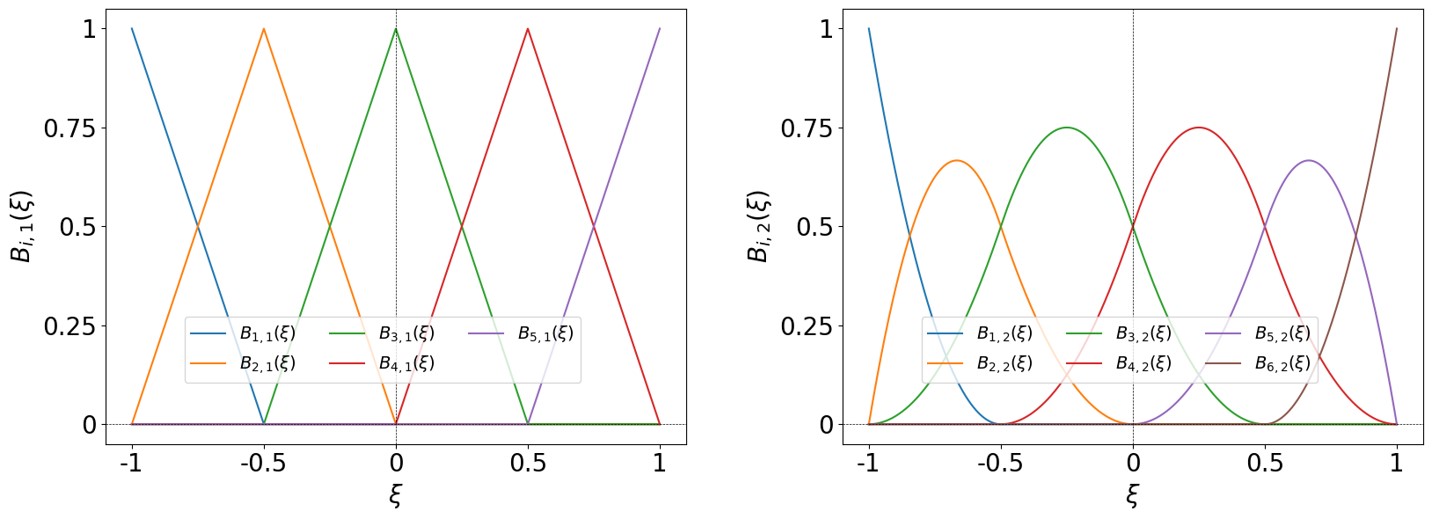}
    \end{center}
    \caption {B-spline (left) $p$=1 $\xi = \{-1, -1, -0.5, 0, 0.5, 1, 1\}$,
    and (right) $p$=2 $\xi = \{-1, -1, -1, -0.5, 0, 0.5, 1, 1, 1\}$.}
    \label{fig14}
    \end{figure*}


\section{Material properties \label{sec:appx2}}
\subsection{Ogden model and Prony series \label{sec:appx2.1}}
Ogden model~\cite{lee2019prediction} is
$
    U = \sum_{i=1}^{N} \frac{2\delta_i}{\alpha_i^2} \left(\lambda_1^{\alpha_i} + \lambda_2^{\alpha_i} + \lambda_3^{\alpha_i} + \lambda_4^{\alpha_i} - 3\right) 
    + \sum_{i=1}^{N} \frac{1}{D_i} \left(U - 1\right)^2,
$ where $\bar{\lambda}_n$ represents the principal stretches, $J$ is the volume ratio, $\delta_i$ and $\alpha_i$ are determined by a fitting with the stress-strain curve of the material, and $D_i$ means the compressibility. Prony series for viscoelastic~\cite{lee2019prediction} is 
$
    G(t) = G_0 \left[ 1 - \sum_{i=1}^{N} g_i e^{\left( \frac{-t}{\tau_i} \right)} \right],
$ where $G(t)$ and $G_0$ are the shear modulus and the initial shear modulus. The terms $g_i$ and $\tau_i$ represent the dimensionless shear-relaxation modulus and the relaxation time. 

\begin{table}[!ht]
    \caption{Material properties of the Ogden model and Prony series~\cite{lee2019prediction} with a density of 1200 $[\mathrm{kg/m^{-3}}]$~\cite{abu2016thermal}.}
    \centering
    \renewcommand{\arraystretch}{1.5} 
    \setlength{\tabcolsep}{4pt}       
    
    {\small 
    \begin{tabular}{p{2.5cm} p{0.5cm} p{2.5cm} p{0.5cm} p{2.5cm} p{0.5cm} p{3.0cm}}
    \hline 
    \textbf{Parameters} & & \textbf{Hyperelasticity (Ogden)} & & \textbf{Parameters} & & \textbf{Viscoelasticity (Prony)} \\
    \hline 
    $\mu_{1}$ (Pa) & & -61.14 & & $g_{1}$ & & 0.26 \\
    $\mu_{2}$ (Pa) & & 25.51 & & $g_{2}$ & & 0.16 \\
    $\mu_{3}$ (Pa) & & 37.42 & & $\tau_{1}$ & & 0.015 \\
    $\alpha_{1}$ & & 3.814 & & $\tau_{2}$ & & 70.3 \\
    $\alpha_{2}$ & & 4.189 & & & & \textbf{Isotropic elasticity} \\ \cline{5-7} 
    $\alpha_{3}$ & & 3.423 & & $\mathrm{E}$ (Pa) & & 0.26 \\
    $D_{1}$ (Pa$^{-1}$) & & 0.1 & & $\nu$ & & 0.16 \\
    $D_{2}$ (Pa$^{-1}$) & & 0.0 & & $\mathit{K}$ (Pa) & & 0.015 \\
    $D_{3}$ (Pa$^{-1}$) & & 0.0 & & $\mathrm{G}$ (Pa) & & 70.3 \\
    \hline 
    \end{tabular}}
    \label{tab:mat8}
\end{table}

\subsection{Structural steel and stainless steel \label{sec:appx2.2}}
\begin{table}[!ht]
    \caption{Material properties of structural steel and stainless steel~\cite{madenci2015finite}.}
    \centering
    \renewcommand{\arraystretch}{1.5} 
    \setlength{\tabcolsep}{4pt}       
    
    {\small 
    \begin{tabular}{p{5cm} p{0.8cm} p{3.5cm} p{0.8cm} p{3.5cm}}
    \hline 
    \textbf{Parameters} & & \textbf{Structural steel} & & \textbf{Stainless steel} \\
    \hline 
    Density (kg/m$^{3}$) & & $7,850$ & & $7,750$ \\
    Young's modulus (Pa) & & $2.00\times10^{11}$ & & $1.93\times10^{11}$ \\
    Shear modulus (Pa) & & $7.6923\times10^{10}$ & & $7.3664\times10^{10}$ \\
    Bulk modulus (Pa) & & $1.6667\times10^{11}$ & & $1.6930\times10^{11}$ \\
    Poisson's ratio  & & $0.30$ & & $0.31$ \\
    \makecell[l]{Isotropic thermal conductivity \\ ($\mathrm{W}\cdot\mathrm{m^{-1}}\mathrm{C^{-1}}$)} & & $60.5$ & & $15.1$ \\
    \makecell[l]{Specific heat constant pressure \\ ($\mathrm{J\cdot kg^{-1}C^{-1}}$)} & & $434.0$ & & $480.0$ \\
    Isotropic relative permeability & & $10,000.0$ & & $1.0$ \\
    Isotropic resistivity ($\Omega\cdot\mathrm{m}$) & & $1.7\times10^{-7}$ & & $7.7\times10^{-7}$ \\
    Tensile yield strength (Pa) & & $2.50\times10^{8}$ & & $2.07\times10^{8}$ \\
    Compressive yield strength (Pa) & & $2.50\times10^{8}$ & & $2.07\times10^{8}$ \\
    Tensile ultimate strength (Pa) & & $4.60\times10^{8}$ & & $5.86\times10^{8}$ \\
    Compressive ultimate strength (Pa) & & $0.00$ & & $0.00$ \\
    \hline 
    \end{tabular}}
    \label{tab:mat9}
\end{table}

\end{appendices}

\newpage
\bibliography{Optimal_SDD}
\bibliographystyle{abbrv}


\end{document}